%% file: main.tex
\documentclass[Afour,sageh,times]{sagej}

\usepackage{moreverb} 
\usepackage{url}
\usepackage[inline,shortlabels]{enumitem}
\usepackage[ruled,vlined,linesnumbered]{algorithm2e}
\usepackage{multirow}
\usepackage[dvipsnames]{xcolor}
\usepackage{tikz}
\usepackage{xspace}
\usepackage{microtype}
\usepackage{caption}
\usepackage[colorlinks,bookmarksopen,bookmarksnumbered,citecolor=red,urlcolor=blue]{hyperref}
\usepackage{doi} 
\usepackage{cleveref}
\usepackage{orcidlink}

\usetikzlibrary{arrows.meta,backgrounds,fit,positioning,shapes.geometric,bending}

\renewcommand{\journalname}{}
\renewcommand{\volumenumber}{}
\renewcommand{\issuenumber}{}
\renewcommand{\volumeyear}{}

\makeatletter
\def\ps@sagepage{%
\let\@mkboth\@gobbletwo
\def\@evenhead{}%
\def\@oddhead{}%
\def\@evenfoot{}%
\def\@oddfoot{}%
}

\def\ps@title{%
\let\@mkboth\@gobbletwo
\def\@evenhead{}%
\def\@oddhead{}%
\def\@evenfoot{\reset@font\hfil\thepage\hfil}%
\def\@oddfoot{\reset@font\hfil\thepage\hfil}%
}

\def\@maketitle{%
\vspace*{-34pt}%
\null%
\begin{center}
\if@PCfour
\begin{rm}
\else
\begin{sf}
\fi
\begin{minipage}[t]{\textwidth}
  \vskip 12.5pt%
    {\raggedright\titlesize\textbf{\@title} \par}%
    \vskip 1.5em%
    \vskip 12.5mm%
\end{minipage}
{\par\large%
      \if@Royal
      \vspace*{6mm}
      \fi
      \if@Crown
      \vspace*{6mm}
      \fi%
      \lineskip .5em%
      {\raggedright\textbf{\@author}
      \par}}
     \vskip 40pt%
    {\noindent\usebox\absbox\par}
    {\vspace{20pt}%
      {\noindent\normalsize\@keywords}\par}
      \if@PCfour
      \end{rm}
      \else
      \end{sf}
      \fi
      \end{center}
      \if@Royal
      \vspace*{-4.5mm}
      \fi
      \if@Crown
      \vspace*{-4.5mm}
      \fi
      \vspace{22pt}
        \par%
  }
\makeatother

\newcommand{\neurolog}{\textsc{NeuroLog}\xspace}

\newcommand{\exemplarSet}{\ensuremath{\widehat{\targetPolicy}}\xspace}
\newcommand{\trainingSet}{\ensuremath{\exemplarSet_{train}}\xspace}
\newcommand{\validationSet}{\ensuremath{\exemplarSet_{val}}\xspace}

\newcommand{\testSet}{\ensuremath{\exemplarSet_{test}}\xspace}

\newcommand{\setOfAtoms}{\ensuremath{A}\xspace}
\newcommand{\setOfTargetPolicies}{\ensuremath{P}\xspace}
\newcommand{\targetPolicy}{\ensuremath{p}\xspace}
\newcommand{\relativeFitness}{\ensuremath{f}\xspace}

\newcommand{\notion}[1]{\emph{\textbf{#1}}}

\newcommand{\symMod}{\texttt{SymbolicModule}\xspace}
\newcommand{\neurMod}{\texttt{NeuralModule}\xspace}
\newcommand{\translator}{\texttt{Translator}\xspace}

\newcommand{\impliesRule}{\,\texttt{implies}\;\,}
\newcommand{\atom}[1]{\texttt{#1}\xspace}
\def\neg{\texttt{-}}
\newcommand{\specialAtom}{\atom{head}}

\newcommand{\mutationAdd}{\ensuremath{S_{+}}\xspace}
\newcommand{\mutationSimplify}{\ensuremath{S_{\downarrow}}\xspace}
\newcommand{\mutationClone}{\ensuremath{S_{0}}\xspace}

\newcommand{\neuralWeights}{\ensuremath{w}\xspace}
\newcommand{\mutationNNWeightsInherited}{\ensuremath{N_{\text{p}\neuralWeights}}\xspace}
\newcommand{\mutationNNWeightsNotInherited}{\ensuremath{N_{\text{r}\neuralWeights}}\xspace}

\newcommand*\mnistOne{\raisebox{-0.17\baselineskip}{\includegraphics[height=0.81\baselineskip]{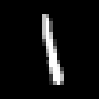}}}
\newcommand*\mnistTwo{\raisebox{-0.17\baselineskip}{\includegraphics[height=0.81\baselineskip]{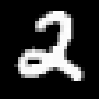}}}

\newcommand{\lineage}{\ensuremath{\ell}\xspace}
\newcommand{\parent}{\ensuremath{parent}\xspace}
\newcommand{\fittest}{\ensuremath{fittest}\xspace}
\newcommand{\generation}{\ensuremath{gen}\xspace}
\newcommand{\maxgen}{\ensuremath{maxgen}\xspace}
\newcommand{\population}{\ensuremath{pop}\xspace}
\newcommand{\githubRepo}{\url{https://github.com/CYENS/evolvable-nesy}}
\begin{document}
\runninghead{Thoma et al.}
\title{Neural-Symbolic Integration with Evolvable Policies}

\author{%
    Marios Thoma\affilnum{1,2}~\orcidlinkc{0000-0001-7364-5799},
    Vassilis Vassiliades\affilnum{1}~\orcidlinkc{0000-0002-1336-5629} and
    Loizos Michael\affilnum{2,1}
}
\affiliation{%
    \affilnum{1} CYENS Centre of Excellence, Nicosia, Cyprus\\
    \affilnum{2} Open University of Cyprus, Nicosia, Cyprus%
}
\corrauth{%
    Marios Thoma,
    Dimarchou Lellou Demetriadi 1,
    Nicosia, 1016,
    Cyprus.
}

\email{m.thoma@cyens.org.cy}

\begin{abstract}
    Neural-Symbolic (NeSy) Artificial Intelligence has emerged as a promising approach for combining the learning capabilities of neural networks with the interpretable reasoning of symbolic systems.
    However, existing NeSy frameworks typically require either predefined symbolic policies or policies that are differentiable, limiting their applicability when domain expertise is unavailable or when policies are inherently non-differentiable.
    We propose a framework that addresses this limitation by enabling the concurrent learning of both non-differentiable symbolic policies and neural network weights through an evolutionary process.
    Our approach casts NeSy systems as organisms in a population that evolve through mutations (both symbolic rule additions and neural weight changes), with fitness-based selection guiding convergence toward hidden target policies.
    The framework extends the \neurolog architecture to make symbolic policies trainable, adapts Valiant's Evolvability framework to the NeSy context, and employs Machine Coaching semantics for mutable symbolic representations.
    Neural networks are trained through abductive reasoning from the symbolic component, eliminating differentiability requirements.
    Through extensive experimentation, we demonstrate that NeSy systems starting with empty policies and random neural weights can successfully approximate hidden non-differentiable target policies, achieving median correct performance approaching 100\%.
    This work represents a step toward enabling NeSy research in domains where the acquisition of symbolic knowledge from experts is challenging or infeasible.
\end{abstract}


\maketitle
%
%
\section{Introduction}

In recent years, Artificial Intelligence (AI) has seen rapid integration into everyday life, with technologies such as computer vision, personalized recommendation systems, facial recognition, and Large Language Models (LLMs) becoming increasingly prevalent.
However, despite the many advantages of Deep Learning and Neural Networks (NNs), their ``black-box'' nature~\citep{liang2021BlackBoxes} presents a significant challenge for developing trustworthy and explainable AI systems.

Neural-Symbolic AI (NeSy)~\citep{besold2021Chapter1,hitzler2021NeSyBook} addresses this challenge by combining the learning capabilities of NNs with the inherent explainability of symbolic AI.
The goal is to develop robust AI systems capable of perception-based learning \emph{and} logical reasoning, thereby enhancing their interpretability, generalizability, and transparency~\citep{garcez2023Neurosymbolic}.
A recent notable example of such integration is DeepMind's AlphaGeometry~\citep{trinh2024AlphaGeometry}, which combines an LLM with a symbolic reasoning engine to solve geometry problems from the International Mathematical Olympiad, achieving scores nearing those of human gold-medalists.

A central challenge in Neural-Symbolic AI concerns the acquisition of \notion{symbolic policies}\endnote{The term ``policy'' is used to refer to \emph{logic-based knowledge}, referred to variously in the literature as ``theory''~\citep{tsamoura2021Neurolog}, ``program''~\citep{gaunt2017Differentiable}, ``knowledge base''~\citep{michael2019MachineCoaching}, among others.} when domain expertise or background knowledge is not available or difficult to obtain.
Existing approaches address this challenge through various strategies: some methods learn differentiable approximations of symbolic policies through gradient-based optimization~\citep[e.g.,][]{dai2019Bridging,serafini2016LogicTensorNets}, while others assume fixed symbolic structures that guide neural learning without imposing differentiability constraints~\citep[e.g.,][]{tsamoura2021Neurolog}.
However, a critical gap remains in the concurrent learning of non-differentiable symbolic policies alongside neural components, without relying on predefined symbolic policies or differentiability assumptions.

In this work, we propose a framework that addresses this gap by enabling NeSy systems to concurrently learn symbolic policies and optimize neural weights through an evolutionary process.
Our key contribution is the demonstration that non-differentiable symbolic policies can be learned from scratch (starting with empty policies), while simultaneously training neural networks, without requiring either predefined knowledge or gradient-based policy learning.
Building on Valiant's Evolvability framework~\citep{valiant2009Evolvability} and broader work on integrating coachable policies with neural architectures~\citep{michael2023Chapter11Autodidactic}, we treat NeSy systems as organisms in an evolutionary population, where mutations introduce new symbolic rules and adjust neural weights, and fitness-based selection guides the system toward approximating hidden target policies.

Through extensive experimentation involving 150 runs across 30 randomly generated target policies, this study shows that NeSy systems starting with empty symbolic policies and randomly initialized neural weights can successfully approximate hidden non-differentiable target policies.
The evolutionary approach reliably achieves a median final correct performance of near 100\% correct performance, establishing its viability for concurrent neural and symbolic learning in settings without predefined knowledge or differentiability assumptions.

Although the approach attains high correctness, it does so at a high computational cost when compared to an end-to-end neural baseline.
The phrase \emph{``price of interpretability/explainability''} has been used to quantify various trade-offs in machine learning, including predictive or reward gaps \citep{bertsimas2019PriceInterpretability,garcia2024InterpretablePoliciesPrice}, degradation of clustering objectives under explainability constraints \citep{dasgupta2020ExplainableKmeansClustering,laber2021PriceExplainabilityCluster}, and differences in financial return under regulatory constraints \citep{dessain2023CostExplainabilityAI}.
In this work, we operationalize the price as additional computational resources and longer training required to obtain explicit, human-readable symbolic policies relative to end-to-end neural baselines that achieve similar predictive performance more efficiently, as reported in \Cref{sec:baselines}.

The remainder of this paper is organized as follows.
We begin by providing the necessary background information and detailing the proposed framework in \Cref{sec:preliminaries,sec:framework}.
We present our empirical evaluation and findings in \Cref{sec:experiments-and-results}, followed by their analysis in \Cref{sec:discussion}.
Finally, we relate our work to the existing literature in \Cref{sec:relatedWork}, before offering conclusions and directions for future work in \Cref{sec:conclusionsFutureWork}.

\section{Background \& Preliminaries} \label{sec:preliminaries}

The framework we propose draws upon three key concepts in the literature to enable the concurrent learning of neural and symbolic components:
\begin{enumerate*}[(i)]
    \item the NeSy framework \neurolog~\citep{tsamoura2021Neurolog},
    \item Valiant's Evolvability framework~\citep{valiant2009Evolvability}, and
    \item Machine Coaching~\citep{michael2019MachineCoaching}.
\end{enumerate*}
We briefly discuss these concepts below, as well as our contributions that make their use together possible (for more details, we direct the reader to the original papers).

\subsection{Machine Coaching} \label{subsec:machineCoaching}

The original \neurolog framework used the Prolog language to represent symbolic policies (see \Cref{subsec:originalNeurolog}), which are not inherently mutable or elaboration tolerant~\citep{mccarthy1998Elaboration}, thus posing challenges for their integration with the evolutionary learning mechanism used in our proposed framework (described in \Cref{subsec:evolvabilityFramework}).
To overcome this limitation, we adopt the semantics of Machine Coaching (MC)~\citep{michael2019MachineCoaching}, an argumentation-based learning framework that structures symbolic knowledge into policies composed of prioritized sets of rules, making them inherently mutable and elaboration tolerant, properties that enable their integration with neural architectures in evolvable systems~\citep{michael2023Chapter11Autodidactic}.

Specifically, following MC's semantics, binary concepts in the universe of discourse are represented using propositional logic, in the form of \notion{atoms}.
\notion{Literals} are instances of atoms, positive (e.g.\ ``\atom{a}'') or negative (e.g.\ ``\neg{\atom{a}}'') (MC adopts an open-world assumption, where the absence of an atom in perception \emph{does not} imply its negation).
In turn, a \notion{rule} is a logical construct composed of a sequence of literals forming its \notion{body}, and a single literal forming its \notion{head}.
If all the literals in a rule's body are verified to be true given a \notion{context} of literals, the rule's head is concluded to be true as well.~E.g.,
rule~``\atom{a},~\atom{b}~\impliesRule\atom{c}'', given the context \(\{\atom{a}, \atom{b}\}\), will conclude that ``\atom{c}'' holds.

\begin{figure}[t]
    \includegraphics[width=0.7\linewidth]{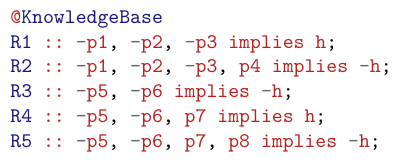}
    \caption{
        Example symbolic policy structure using Machine Coaching semantics.
        The policy consists of prioritized rules (increasing in priority from top to bottom), where each rule has a body (sequence of literals) and head (single literal).
        This example demonstrates the type of non-differentiable shallow propositional policies that our evolutionary framework learns to approximate.
        This policy is used throughout the paper to demonstrate the framework's operation.
    }
    \label{fig:kbExample}
\end{figure}

A \notion{policy} organizes a set of rules into a prioritized sequence, with later rules having higher priority.
Given a context of literals, a policy's decision is determined by the highest-priority rule that is triggered by the context.
This decision-making process can yield \notion{correct} or \notion{wrong} predictions, but also \notion{abstentions}, when no rule is triggered by the context.
An example of such a policy structure is shown in \Cref{fig:kbExample}.

MC allows new rules to be added to the end of a policy, where they automatically have higher priority than existing rules and can override them.
This property makes rule additions suitable as \emph{symbolic mutations} in our evolutionary framework.

Although rules in a policy can form chains, where the conclusion of one rule serves as input to subsequent rules, in this study we restrict our attention to \emph{shallow propositional policies}, i.e. policies with no chained rules, where all rules have the special atom \specialAtom (or \neg{\specialAtom}) as their head.

\subsection{Evolvability Framework} \label{subsec:evolvabilityFramework}

Valiant's \emph{Evolvability} framework~\citep{valiant2009Evolvability}, shown to be a restricted case of Probably Approximately Correct (PAC) learnability~\citep{valiantPAC1984}, formalizes evolution as a learning mechanism seeking to approximate a hidden target function.
It does so by casting functions as \notion{organisms} in a population, evolving through mutation and survival-of-the-fittest, to gradually approximate the target function.
Adapting this to a NeSy context, we treat NeSy systems as \notion{NeSy organisms}.

In the original framework, each organism in the population is a mutated \notion{offspring} derived from a \notion{parent} organism of the preceding generation.
To accommodate the dual nature of NeSy organisms, we implement mutation mechanisms for both neural and symbolic components:
\begin{enumerate*}[(i)]
    \item \notion{neural mutations}, which involve the offspring inheriting the neural network weights from their parent, or starting with randomly initialized weights; and
    \item \notion{symbolic mutations}, which involve the offspring's symbolic policy being expanded by adding a new rule, building upon the \parent's existing policy, thus incrementally evolving the organism's symbolic reasoning capabilities.
\end{enumerate*}

Evolution proceeds in distinct generations.
At the start of each generation, the \parent organism reproduces to create a population of offspring (the \parent itself is not part of the generation), which due to the mutations is variable.
Each organism in the population is evaluated for \notion{fitness} against a dataset created using the hidden target function.
We adopt the same mechanism, with the only difference being that NeSy organisms are first trained on a \emph{training} dataset labeled using a hidden target \emph{symbolic policy}, then evaluated for fitness using a \emph{validation} dataset.
We employ a \notion{relative fitness} metric that compares each offspring's performance against the validation dataset to their \parent's performance.

The selection of the \fittest organism follows the mechanism of the original framework: offspring are grouped according to the relation of their relative fitness to a fixed threshold parameter $t$.
An organism $o_i$ is categorized as \notion{detrimental}, \notion{neutral}, or \notion{beneficial}, if their relative fitness $\relativeFitness_i$ is in $(-\infty,-t)$, $[-t,+t]$, or $(+t,+\infty)$, respectively.
If available, a beneficial offspring $o_i$ is selected using a form of fitness proportionate selection, with probability $\relativeFitness_i^k / \sum_j \relativeFitness_j^k$, where the exponent $k$ is a fixed nonlinearity parameter.
If beneficial offspring are not available, a neutral offspring $o_i$ is selected uniformly at random.
Finally, if the other two groups are empty\endnote{The original \emph{Evolvability} framework guarantees that the neutral group will be nonempty by means of clone mutations (Sec.~\ref{sec:evol-process}), but in our case this is \emph{not} guaranteed, due to the randomness introduced by the \neurMod and its interaction with the \symMod.}, the offspring $o_i$ with the highest $\relativeFitness_i$ from the detrimental group is selected, breaking any ties at random.
The selected organism is added to the \notion{lineage} of \fittest organisms and becomes the founder of the next generation.

Past work by~\citet{markos2022ProxyCoaches} demonstrated the effectiveness of this evolutionary framework for learning symbolic policies in conjunction with Machine Coaching semantics.
Building on this foundation, our work extends the approach to the neural-symbolic setting by incorporating the \neurolog architecture described below, enabling the concurrent evolution of both symbolic policies and neural network weights.


\subsection{\neurolog NeSy Framework} \label{subsec:originalNeurolog}

The \neurolog framework~\citep{tsamoura2021Neurolog} compositionally integrates neural and symbolic systems by treating them as ``black-boxes'' incorporated as independent \notion{modules} in a unified architecture.
The architecture consists of a \symMod (containing a symbolic policy), a \neurMod (containing a NN), and a predefined \translator function that enables communication between the modules (\Cref{fig:neurologArchitecture}).
The framework's compositional nature and semantic agnosticism allowed us to easily adapt it to our proposed evolutionary approach.

The \neurolog framework assumes that the symbolic policy is given, but makes no assumptions about its differentiability.
It bypasses differentiability requirements through \emph{abductive reasoning}~\citep{kakas2017Abduction}, a form of backward reasoning that enables training the \neurMod's neural network using the \symMod's symbolic policy.
A policy \targetPolicy consists of logical expressions over a set of atoms \setOfAtoms (binary/boolean concepts).
The policy supports both forward and backward reasoning: given input atoms $I \subseteq \setOfAtoms$, forward reasoning produces output atoms $O =$ \texttt{deduce}$(\targetPolicy, I)$ such that $\targetPolicy \cup I \models O$.
Conversely, given output atoms $O$, backward reasoning produces one or more sets of input atoms $I \in$ \texttt{abduce}$(\targetPolicy, O)$ such that $\targetPolicy \cup I \models O$.
Any conjunction of atoms from \texttt{abduce}$(\targetPolicy, O)$ forms an \notion{abductive proof} for the given outcome (or label) $O$.

Atoms in \setOfAtoms correspond to both the \neurMod's output neurons and the \symMod's binary concepts (further discussed in \Cref{subsec:machineCoaching}).
During training, the \neurMod learns via abductive feedback from the \symMod through semantic loss calculation~\citep{xu2018SemanticLoss}.
Given a training instance with neural input and its label, the system first generates all possible abductive proofs for the label.
These proofs are compiled into a Sentential Decision Diagram (SDD)~\citep{darwiche2011SDD}, which enables efficient Weighted Model Count (WMC) calculation~\citep{chavira2008WMC}.
The WMC incorporates the neuron activation values produced when the \neurMod processes the neural input, weighting each atom in the abductive proofs.
The semantic loss is then computed as the negative logarithm of the WMC.

\begin{figure}[t]
    \centering
    \resizebox{1\linewidth}{!}{%
        \input{assets/figures/neurologTikz}
    }
    \caption{%
        The \neurolog NeSy architecture, as proposed by~\protect\citet{tsamoura2021Neurolog}.
    }
    \label{fig:neurologArchitecture}
\end{figure}

\begin{figure*}[t]
    \centering
    \includegraphics[width=0.7\textwidth]{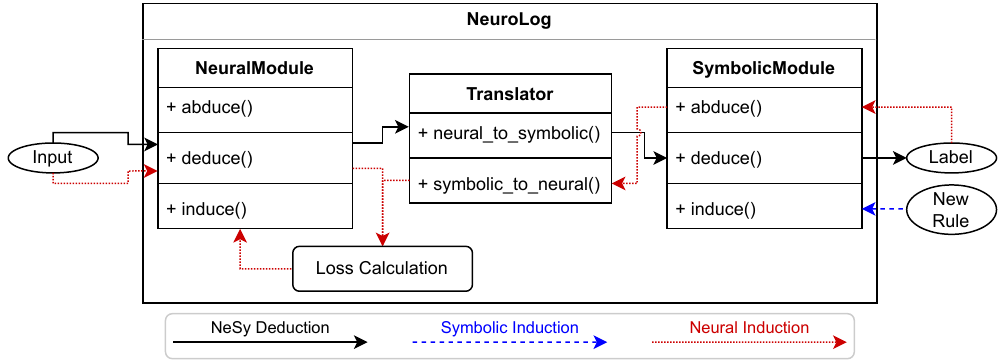}
    \caption{%
        The extended \neurolog architecture enabling concurrent neural and symbolic learning.
        Black arrows show forward reasoning (deduction): the \texttt{deduce()} methods of \neurMod and \symMod are chained via the \translator to produce predictions.
        Red arrows show neural induction: the \neurMod trains via backpropagation using abductive feedback from the \symMod's \texttt{abduce()} method, which generates training signals from symbolic rules.
        Blue arrows show symbolic induction: new rules are added to the \symMod's policy via its \texttt{induce()} method during evolutionary mutations.
    }
    \label{fig:neurolog-all}
\end{figure*}

\section{Proposed Framework} \label{sec:framework}

Bringing together the ideas discussed in the previous section, our proposed framework consists of two major components: an extensively modified and elaborated \neurolog architecture, and an evolutionary algorithm, which, when combined, allow the concurrent training of both neural and symbolic components of the NeSy system.

\subsection{Extended \neurolog Architecture} \label{subsec:extended-neurolog}

We further formalize the \neurolog framework to introduce a learning mechanism to the \symMod, to make it trainable in an evolutionary setting.

The original framework employs the notions of \emph{deduction} (forward reasoning) and \emph{abduction} (backward reasoning), to allow for the transparent integration between the neural and symbolic modules.
We expand this to include the notion of \emph{induction} (learning), i.e.\ a way for both modules to be trainable, which was already possible for the \neurMod,
but not for the \symMod.
We do so by requiring both modules to expose three methods: \texttt{deduce()}, \texttt{abduce()} and \texttt{induce()}, for the three notions mentioned above, respectively\endnote{It should be noted that the \texttt{abduce()} method does not map to any meaningful functionality in the \neurMod, but was included for symmetry's sake.}.

\begin{figure*}[t]
    \centering
    \includegraphics[width=0.97\linewidth]{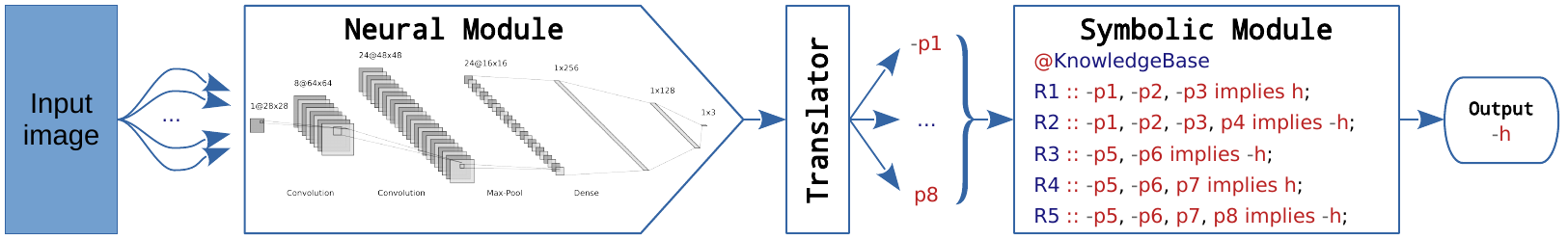}
    \caption{%
        Example of forward reasoning in the extended \neurolog architecture with a concrete target policy.
        The input MNIST image sequence is processed by the \neurMod (CNN), producing symbolic atom predictions via the \translator, which the \symMod uses with its policy (the same policy shown in \Cref{fig:kbExample}) to generate the final output.
        This policy represents one of the randomly generated target policies used in the experiments.
    }
    \label{fig:neurolog-deduction}
\end{figure*}

Forward reasoning in the system is achieved by chaining the \texttt{deduce()} methods of the two modules, using the \translator to facilitate their communication, as demonstrated in \Cref{fig:neurolog-all} (black arrows).
A concrete example of this forward reasoning process with actual MNIST input and a symbolic policy is shown later in \Cref{fig:neurolog-deduction}.

For the \symMod, the \texttt{induce()} method provides a way to modify the symbolic policy in the module, so that new knowledge is acquired.
Although our framework does not make any assumptions about the symbolic policy semantics used, in this study we adopt the MC semantics, as discussed in \Cref{subsec:machineCoaching}.
Thus, the \symMod \texttt{induce()} method maps to the addition of new rules to a symbolic policy (\Cref{fig:neurolog-all}, blue line).
We employ the inference engine \emph{Prudens}~\citep{markos2022Prudens}, which implements MC, to interpret the symbolic policies.
\emph{Prudens} is capable of both \textit{deduction} and \textit{abduction} with such policies, and thus powers both \texttt{deduce()} and \texttt{abduce()} methods of the \symMod.
Since the abduction process is computationally costly, the proposed framework implements extensive caching of abductive results.

In \neurMod, the \texttt{induce()} method directly maps to the use of backpropagation in NNs, once training loss has been computed using abductive feedback from the \symMod \texttt{abduce()} method (\Cref{fig:neurolog-all}, red line).
Training of the system, similar to classical NN training, occurs across multiple epochs, with the full training dataset used in each epoch.

As explained previously, instances of the extended \neurolog architecture are cast as organisms in an evolutionary setting.
Although training of an organism's \neurMod is done primarily through abductive feedback from the \symMod, as explained in \Cref{subsec:originalNeurolog}, when the organism is still learning its symbolic policy, said abductive feedback is incomplete (especially early in the evolutionary process).
To investigate whether additional training signals could improve convergence during these early generations, a self-supervised learning mechanism was introduced to the \neurMod in the form of reconstruction loss.
This auxiliary loss component provides a training signal that is independent of the symbolic policy's completeness, potentially helping guide neural learning when abductive feedback is limited.
The details of the implementation and the results of the ablation study for different loss configurations are presented in \Cref{sec:experimental-setup,sec:ablation-loss-ratios}.

\subsection{Evolutionary Process} \label{sec:evol-process}

The goal of the evolutionary process is to approximate a NeSy system containing a hidden target policy \targetPolicy, starting with a NeSy system with an empty policy, while concurrently optimizing its neural weights.
To do so, an exemplar set \exemplarSet is constructed using the target policy \targetPolicy, and is then split randomly into training (\trainingSet), validation (\validationSet), and testing (\testSet) subsets.
\trainingSet and \validationSet are used during the evolutionary process, while \testSet is held out and used to objectively evaluate the evolutionary process once it is finished.

\Cref{alg:evolutionaryAlgo} outlines the evolutionary process, which uses instances of our modified \neurolog architecture as NeSy organisms.
At $\generation=0$, an array~\lineage, intended to hold the lineage of \fittest organisms, is seeded with a single organism with an empty symbolic policy and randomly initialized neural weights.
Each subsequent generation starts with the \fittest organism of the previous generation reproducing to create a population ($\population_{\generation}$) of mutated offspring.

Specifically, for the symbolic mutations, all offspring inherit the symbolic policy of their \parent, modified as follows:
\begin{enumerate*}[(i)]
    \item \mutationClone (\notion{clone} mutation): an offspring inherits an exact copy of the symbolic policy of its \parent;
    \item \mutationAdd (\notion{addition} mutation): based on a random context $x$ drawn from $2^\setOfAtoms$ (since each atom can be positive or negative), two offspring are created by adding the new rule ``$x$ \impliesRule \specialAtom'' or ``$x$ \impliesRule \neg{\specialAtom}'' to their policy;
    \item \mutationSimplify (\notion{simplification} mutation): an offspring is created for each $j$, by adding the rule ``\atom{body$_{-j}$} \impliesRule \atom{head}'' to their policy, where ``\atom{body} \impliesRule \atom{head}'' is the latest rule added to their \parent's policy, and \atom{body$_{-j}$} is \atom{body} minus its $j$-th literal.
\end{enumerate*}
Similarly, we consider the inheritance or not of the neural weights~\neuralWeights from parent to offspring as neural mutations, specifically:
\begin{enumerate*}[(i)]
    \item \mutationNNWeightsInherited: an offspring that inherits~\neuralWeights from their parent;
    \item \mutationNNWeightsNotInherited: an offspring that starts with randomly initialized~\neuralWeights.
\end{enumerate*}
Enough offspring are created to accommodate all possible combinations of symbolic and neural mutations (the two types of mutation are independent; thus all their combinations are possible, e.g.\ an \(\mutationNNWeightsInherited\mutationClone \) offspring is a symbolic clone with inherited neural weights, while an \(\mutationNNWeightsNotInherited\mutationAdd\) offspring is a symbolic addition mutation with randomly initialized neural weights).

Subsequently, all organisms in $\population_{\generation}$ are trained using \trainingSet for a number of epochs.
Then, their relative fitness is calculated, which is a measure of progress for each offspring from its parent, calculated by a point-by-point comparison of the change in deductions of the offspring vs the parent on \validationSet, rewarding or giving penalties depending on the direction of change, using the score matrix in \Cref{tab:rel_fitness_matrix} (Appendix~\ref{appendix:empirical-setup-detailed}).
Then, based on the parameters $t$ and $k$, a single organism is selected as the \fittest in $\population_{\generation}$, using the mechanism described in \Cref{subsec:evolvabilityFramework}, and added to the lineage \lineage, to become the founder of the next generation.
The evolutionary process continues until
\begin{enumerate*}[(i)]
    \item a specific number of generations is exceeded (\maxgen); or
    \item an early stopping criterion is triggered (e.g.,\ a threshold of correct performance on \validationSet is reached).
\end{enumerate*}
Once the evolutionary process is over, the organisms in \lineage are evaluated using \testSet.
The final organism in~\lineage represents the result of the concurrent training of the neural and symbolic components of a NeSy system, the goal of our proposed framework.

\begin{algorithm}[t]
    \caption{The evolutionary process algorithm.}
    \label{alg:evolutionaryAlgo}
    \SetKwInput{Input}{input}
    \SetKwFor{For}{for}{do}{end}
    \SetKwIF{If}{ElseIf}{Else}{if}{then}{else if}{else}{end if}
    \SetKw{Break}{break}
    \Input{Datasets \trainingSet,  \validationSet \& \testSet. Parameters $t$, $k$ \& \maxgen.}
    At $gen=0$, initialize array \lineage to hold lineage of fittest organisms, with a single organism with empty policy and randomly initialized NN weights\;

    \For{$\generation=1$ \KwTo \maxgen}{
        $\parent=\lineage[gen-1]$\tcc*{Parent fittest organism of previous gen}
        Populate array $\population_{\generation}$ with offspring of \parent with all combinations of neural~(\mutationNNWeightsInherited,~\mutationNNWeightsNotInherited) and symbolic (\mutationClone, \mutationAdd, \mutationSimplify) mutations\;
        Train all organisms in $\population_{\generation}$ using \trainingSet\;
        Using \validationSet, compute relative fitness $\relativeFitness_i$ of each organism $o_i$ in $\population_{\generation}$ vs \parent\;

        Based on $t$, split $\population_{\generation}$ into $beneficial$, $neutral$, and $detrimental$ groups\;
        \If{$beneficial$ is not empty}{
            Select \fittest from $beneficial$\ using fitness proportionate selection with exponent~$k$;
        }
        \ElseIf{$neutral$ is not empty}{
            Select \fittest from $neutral$ uniformly at random\;
        }
        \Else{
            Select \fittest as the organism with highest $\relativeFitness_i$ in $detrimental$\;
        }
        $\lineage[gen]=\fittest$\;
        Clear $\population_{\generation}$\;
        \If{$\relativeFitness_{\fittest} \geq 99\%$}{
            \Break\tcc*[f]{Early stopping}
        }
    }
    Test all organisms in \lineage using \testSet\;
\end{algorithm}

\subsection{Technical Contributions} \label{sec:technical-contribs}

The use of an evolutionary algorithm as a coach involves the repeated mutating (coaching) and training of several NeSy ``organisms'' in successive generations, until a NeSy organism that successfully approximates a hidden symbolic policy emerges.
Preliminary implementations revealed that this process presented significant computational challenges due to two major bottlenecks: the sequential nature of training NeSy organisms during experiments, and the computational cost of loss calculation during training.
Without optimization, training the populations of organisms required for even a single generation was substantially slower than practical, making the experiments described in this study challenging to execute at scale.
The following subsections detail the technical contributions we developed to address these computational bottlenecks.
The effectiveness of these optimizations, which collectively enabled the large-scale experimentation presented in \Cref{sec:experimental-setup}, is empirically validated in \Cref{sec:perf-opt-validation}.

\subsubsection{Parallel Training of Neural Networks on a Single GPU}

A common challenge in training neural networks (NNs) is the VRAM capacity limitation of a single GPU, often necessitating the use of GPUs with larger VRAM, or the adoption of multi-GPU training strategies.
Conversely, the NeSy evolutionary experiments presented in this study involve the training of numerous, relatively small NNs, each embedded within a NeSy organism.
Sequential training of these NNs presents significant computational challenges due to the large number of organisms.
To address this, we developed a parallel training methodology enabling the concurrent training of multiple NNs on a single GPU.
Parallel training was implemented using PyTorch's \texttt{multiprocessing} module.
A significant engineering consideration was ensuring the serializability of all inter-process objects, a requisite of the Python programming environment, so they could be safely passed to other processes.

\subsubsection{Operation Caching During Semantic Loss Calculation} \label{sec:semantic-operation-caching}

The training of NeSy organisms, and particularly their \neurMod, is based on the calculation of semantic loss~\citep{xu2018SemanticLoss}, a process that can be computationally intensive.
This intensity stems from the intricate steps involved in deriving the loss from the abductive feedback of the symbolic component.
As described in \Cref{subsec:originalNeurolog}, for each training instance the \symMod generates abductive proofs, compiles them into a Sentential Decision Diagram (SDD)~\citep{darwiche2011SDD}, and computes the Weighted Model Count (WMC)~\citep{chavira2008WMC} by traversing the SDD through Depth-First Search (DFS) while integrating neural activation values.
The repeated generation of abductive proofs and, more critically, the construction and traversal of SDDs for every batch of training data, represent significant computational overhead.

To mitigate this computational bottleneck, our framework implements an operation caching strategy.
The core idea is to take advantage of the repetitive nature of calculations within and across training batches.
During the first encounter with a specific set of abductive proofs (and thus a specific SDD structure) for a given label within a training batch, the initial DFS operation required for WMC calculation is performed.
The sequence of operations and the structure of this traversal are then cached as a computational graph.
For all subsequent instances within the same batch that require the WMC for the same SDD structure (but typically with different neural activation values), this pre-compiled computational graph is reused.

This caching mechanism substantially reduces training times.
A key advantage is the significant reduction in CPU-GPU interaction.
Traditionally, the symbolic reasoning steps (abduction, SDD construction, DFS) would reside on the CPU, requiring data to be passed back and forth to the GPU where neural computations and loss backpropagation occur.
By caching the computational graph of the WMC calculation, which can be parameterized by the neural outputs, the core of the semantic loss computation can be executed directly and repeatedly on the GPU.
This minimizes the costly data transfers and synchronization points between the CPU and GPU, leading to a more streamlined and efficient use of GPU resources.

\section{Experiments \& Results} \label{sec:experiments-and-results}

This section includes key aspects of our empirical investigation of the proposed framework.
Additional details are included in Appendix~\ref{appendix:empirical-setup-detailed}; code for reproducing all experiments is publicly available (see Supplementary Materials).

\subsection{Target Policies \& Datasets}

The target policies used in the experiments were randomly generated, first by fixing \setOfAtoms to a set of 8 binary concepts $\{a1,\dots,a8\}$ for use in contexts and rule bodies, and then randomly generating a set \setOfTargetPolicies of target policies, using atoms from \setOfAtoms (an example target policy is shown in \Cref{fig:kbExample}).
The random policy generator developed by~\cite{markos2022ProxyCoaches} was used.
Importantly, the target policies used are non-differentiable due to their discrete nature and the logical implications that define their structure, which does not allow for gradient-based optimization methods to be applied directly.

\begin{figure}[t]
    \centering
    \includegraphics[width=1\linewidth]{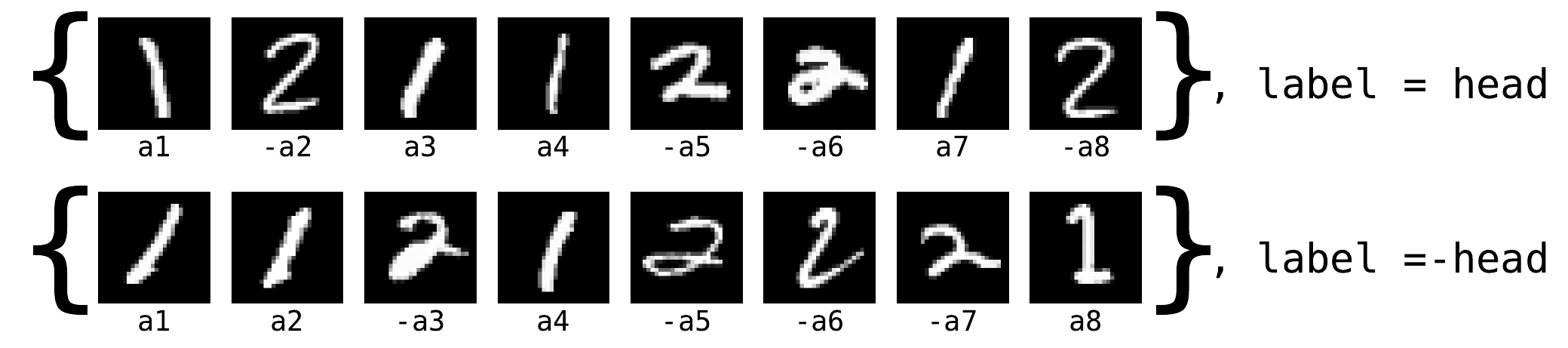}
    \caption{
        Two training instances from a \trainingSet dataset labeled by the symbolic policy in \Cref{fig:kbExample}.
        Each instance consists of a sequence of 8 MNIST digit images representing atoms: digits with value \(1\)~(\mnistOne) represent positive atoms, digits with value \(2\)~(\mnistTwo) represent negative atoms.
        The label (\specialAtom or \neg{\specialAtom}) is determined by applying the symbolic policy's rules to the atom sequence.
    }
    \label{fig:trainingExample}
\end{figure}

Datasets were constructed as follows: to create an exemplar set \exemplarSet for each target policy \(\targetPolicy \in \setOfTargetPolicies\),  data instances (or contexts) consisting of 8 atoms each were constructed by sampling uniformly at random from \(2^\setOfAtoms\)
(since each atom can be positive, negative, or unobserved, but unobserved atoms were not included in the data instances), and labeling each data instance according to \targetPolicy, and filtering out those on which \targetPolicy abstained.
Each training instance was then converted to a pictorial representation, by using randomly picked images of handwritten digits from the MNIST dataset~\citep{lecun1998GradientBased} to represent their individual atoms, using the following convention: MNIST digits with a numerical value of 1~(\mnistOne) were used to represent positive atoms and 2~(\mnistTwo) negative atoms.
Thus, for example, the context \(\{\atom{a1}, \dots, \atom{\neg{a8}}\}\) would be represented as \(\{\mnistOne, \dots, \mnistTwo\}\).
\Cref{fig:trainingExample} shows two examples of full-context training data instances, labeled using the policy in \Cref{fig:kbExample}.
\trainingSet and \validationSet used images from the train subset of MNIST, while \testSet images from the test MNIST subset.

\begin{figure*}[t]
    \centering
    \includegraphics[width=1\linewidth]{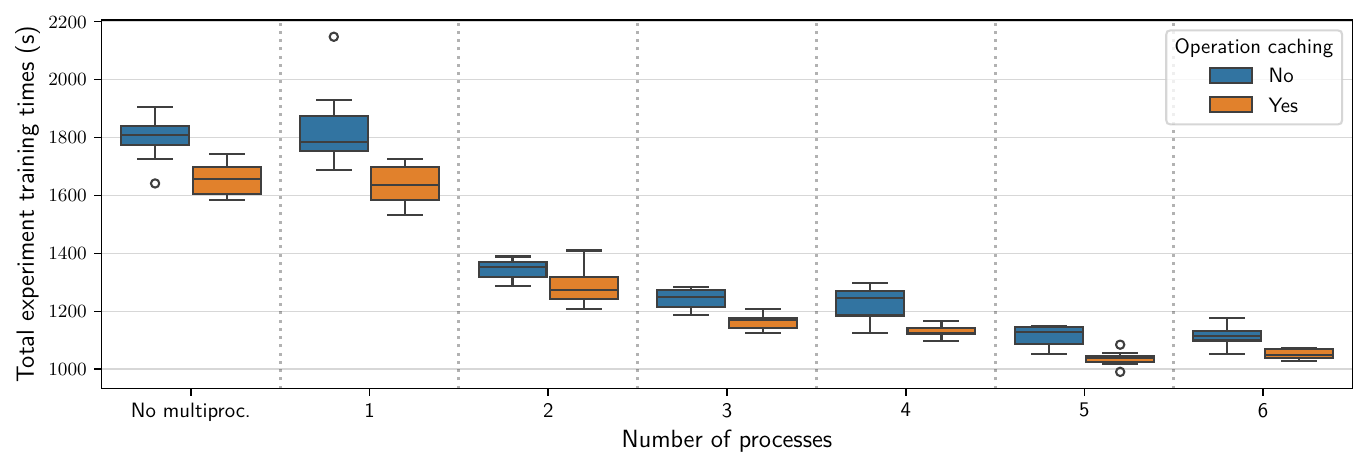}
    \caption{%
        Box plots showing the impact of operation caching and parallel processing on total experiment training times, in experiments involving the training of multiple NeSy organisms.
        Each box represents 20 experiments (10 symbolic policies, each run twice), with each experiment involving the training of 100 NeSy organisms (in a non evolutionary setting).
        The x-axis indicates the number of parallel processes used (where each process corresponds to one CPU) on a single GPU, while the y-axis shows the total experiment training time in seconds.
        Blue boxes represent training without operation caching in semantic loss calculation, while orange boxes show training with operation caching enabled.
        Results demonstrate that operation caching consistently reduces training time across all parallelization configurations.
        The ``No multiproc.'' condition represents sequential training using a single CPU without the multiprocessing framework, while 1 process represents using a single process within the multiprocessing environment.
        The similarity between these configurations confirms that the multiprocessing environment itself introduces minimal overhead.
        Additionally, increasing the number of parallel processes substantially decreases total experiment duration, with efficiency gains plateauing at approximately 5 parallel processes.
    }
    \label{fig:parallel-gpu-training-operation-caching}
\end{figure*}

\subsection{Experimental Setup} \label{sec:experimental-setup}

For the experiments, a set \setOfTargetPolicies of 30 randomly generated target policies was used.
For each policy, the experiments were repeated 5 times using different randomly generated datasets (\(30 \times 5 = 150\) total experiments).
The datasets had the following sizes: \trainingSet $20 000$ data instances, \validationSet $2 000$, and \testSet $2 000$.
The population of each generation was generated using all possible combinations of symbolic (\mutationClone, \mutationAdd, \mutationSimplify) and neural (\mutationNNWeightsInherited, \mutationNNWeightsNotInherited) mutations.
Experimentation showed that the evolutionary process could be considerably shortened by introducing multiple \mutationAdd mutations per generation, so 5~\mutationAdd mutations were used per generation.

A convolutional neural network (CNN) architecture was used in {\neurMod}s, composed of 2 convolutional layers, followed by 3 fully-connected layers (\Cref{fig:neurolog-deduction,fig:nn-arch}).
Since the data instances consisted of sequences of 8 MNIST images, each image was passed through the CNN sequentially, and the output vectors for each were concatenated into a single output vector (the same base encoder was used for the end-to-end baseline described in \Cref{sec:baselines}, with additional output layers for direct binary classification).
The neural weights of \mutationNNWeightsNotInherited offspring were randomly initialized using Xavier initialization~\citep{glorot2010XavierInit}.
The Adam optimizer~\citep{kingma2017AdamOptimizer} was used, with a learning rate of \(0.001\).

The semantic loss was calculated following the methodology in \Cref{subsec:originalNeurolog}, incorporating the optimization techniques we detail in \Cref{sec:semantic-operation-caching}.
The weighted model counting (WMC) was performed using the PySDD library~\citep{wannesm2024PySDD}.

As discussed in \Cref{subsec:extended-neurolog}, to investigate whether additional training signals could improve convergence when abductive feedback is limited, a reconstruction loss component was incorporated.
For this reconstruction loss, a mirroring de-convolution decoder was used to reconstruct the input MNIST images using the NN output vector as input, employing the mean squared error (MSE) for its calculation.
To constrain the decoder to reconstruct only one version of each MNIST digit (numerical values \(\{1, 2\}\)), a Gumbel-Softmax operation~\citep{jang2017Categorical,maddison2017Concrete} was applied to the NN output vector before being given to the decoder.
To evaluate different approaches to combining these loss components, ablation experiments were conducted comparing three semantic:reconstruction loss ratios: 1:0 (semantic loss only), 1:1 (equal weighting), and 1:10000 (heavily weighted reconstruction loss), as detailed in \Cref{sec:ablation-loss-ratios}.

\begin{figure*}[ht]
    \centering
    \includegraphics[width=1\linewidth]{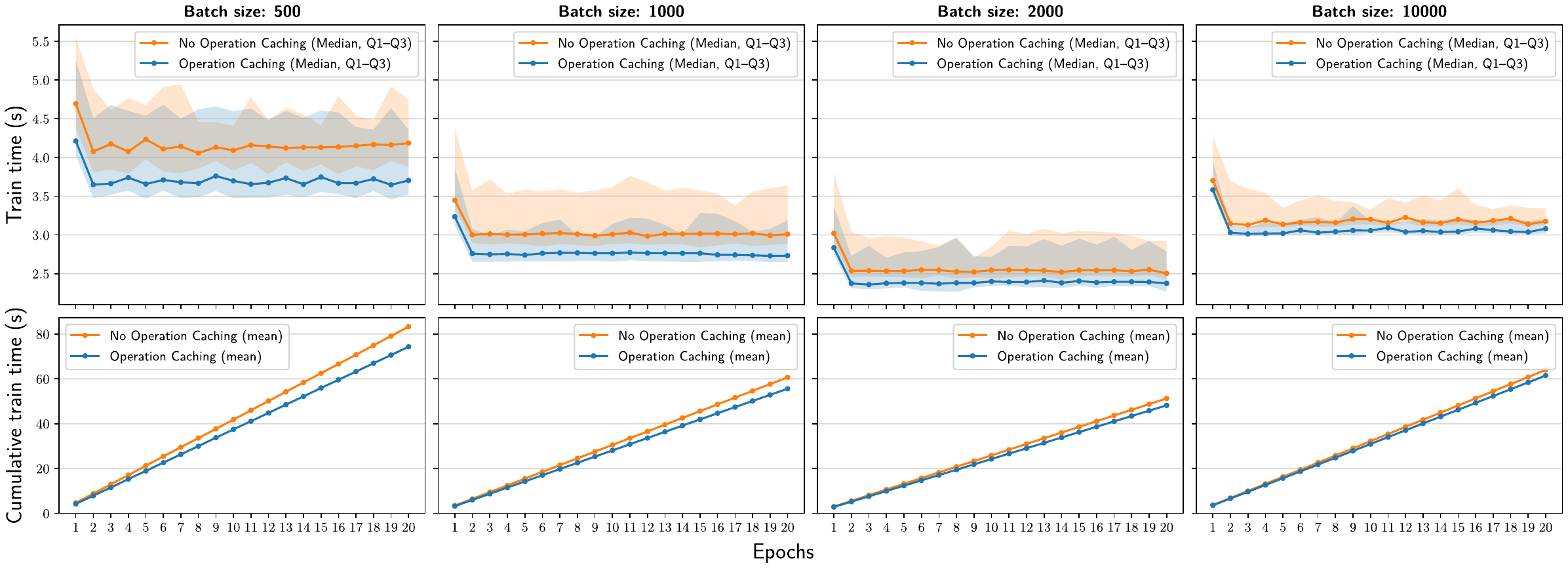}
    \caption{%
        Training time comparison across different batch sizes (500, 1000, 2000, 10000) with and without operation caching during semantic loss calculation.
        For each batch size, the top subplot shows training time per epoch (with median and Q1--Q3 ranges indicated by solid lines and shaded areas, respectively), while the bottom subplot shows cumulative training time (mean) across epochs.
        Batch size 2000 achieves the best overall balance, providing the lowest training time per epoch while maintaining efficient cumulative training time, making it the optimal choice for the main experiments.
    }
    \label{fig:fig-batch-size}
\end{figure*}

For the evolutionary process, the upper limit $\maxgen= 500$ was used, and the early stopping criterion was set for when the correct performance on \validationSet was $\geq 99\%$.
The threshold parameter \(t=0\)  and the non-linearity parameter \(k=2\) were used.
The organisms were trained for 5 epochs using~\trainingSet, with a batch size of 2000.
As shown in \Cref{fig:fig-batch-size}, this batch size was selected because it achieves the best overall balance between training time per epoch and cumulative training efficiency, providing significant performance gains over smaller batch sizes while avoiding the diminishing returns observed at larger batch sizes.

\subsection{Main Results} \label{sec:main-results}

\begin{figure*}[t]
    \centering
    \includegraphics[width=\linewidth]{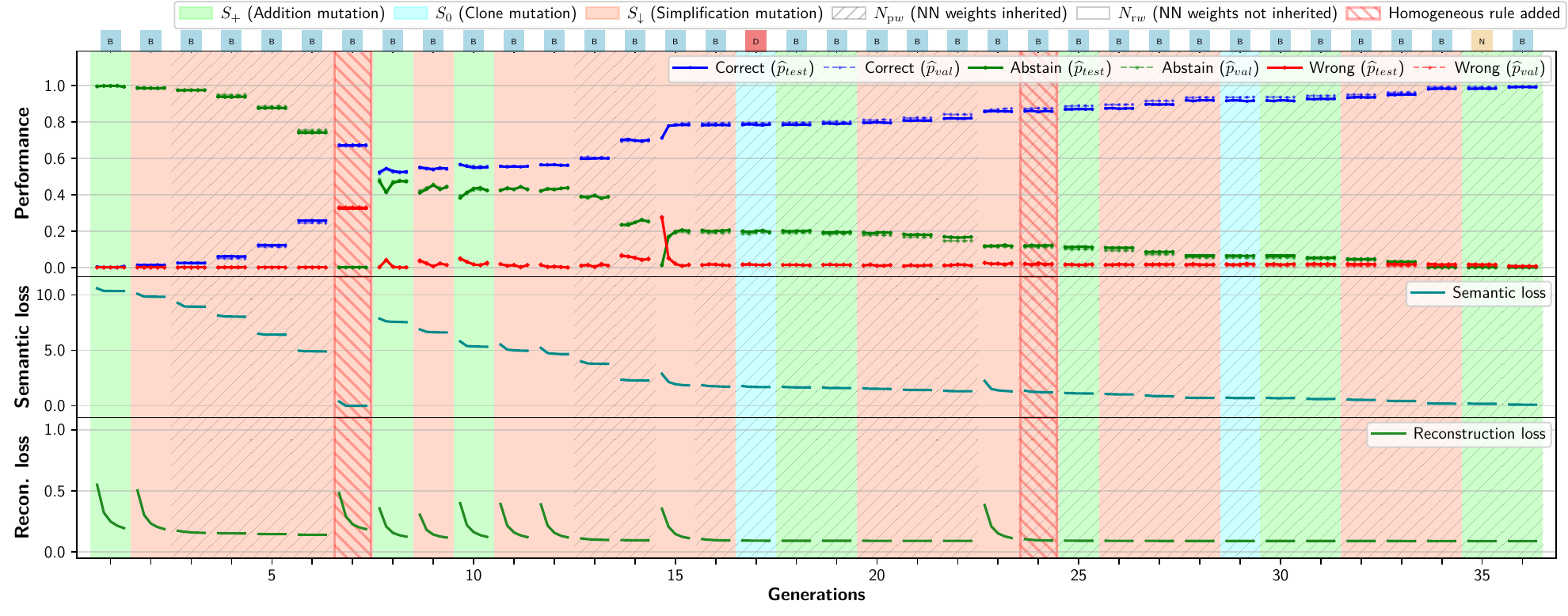}
    \caption{
        Results of a single experimental run: columns represent generations, showing the \fittest organism's performance and training history per generation.
        The 1st subplot shows correct, abstain, and wrong performances on \validationSet and \testSet.
        The 2nd and 3rd subplots show the semantic and reconstruction losses, respectively.
        Symbolic and neural mutations are differentiated by background colors and hatch patterns, with homogeneous rule additions also marked by a hatch pattern.
        The letters on the top x-axis indicate the fitness group the generation's \fittest organism was selected from (beneficial, neutral, or detrimental groups).
        Generation 0 is omitted, since the initial organism is not trained due to an empty policy.
    }
    \label{fig:exp-1-normal}
\end{figure*}

\begin{figure*}[t]
    \centering
    \includegraphics[width=\linewidth]{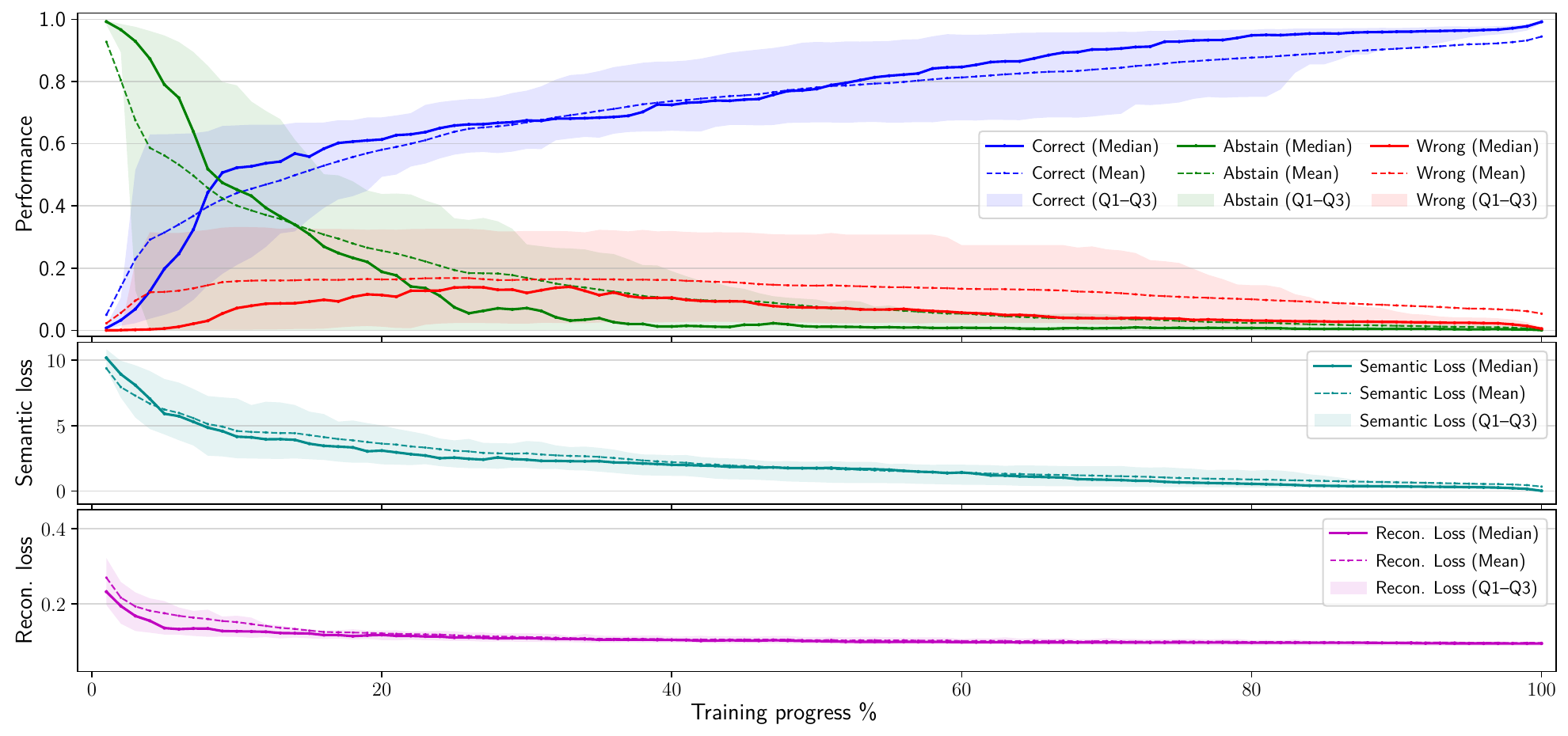}
    \caption{
        Aggregated results of all experiments on their respective \testSet subsets.
        Since experiments concluded at different generation numbers, results were interpolated to a unified 1--100 scale to enable meaningful aggregation across all runs.
        Shaded areas depict the interquartile range (Q1--Q3) for each metric.
    }
    \label{fig:aggregated-results}
\end{figure*}

\begin{figure*}[t]
    \centering
    \includegraphics[width=\linewidth]{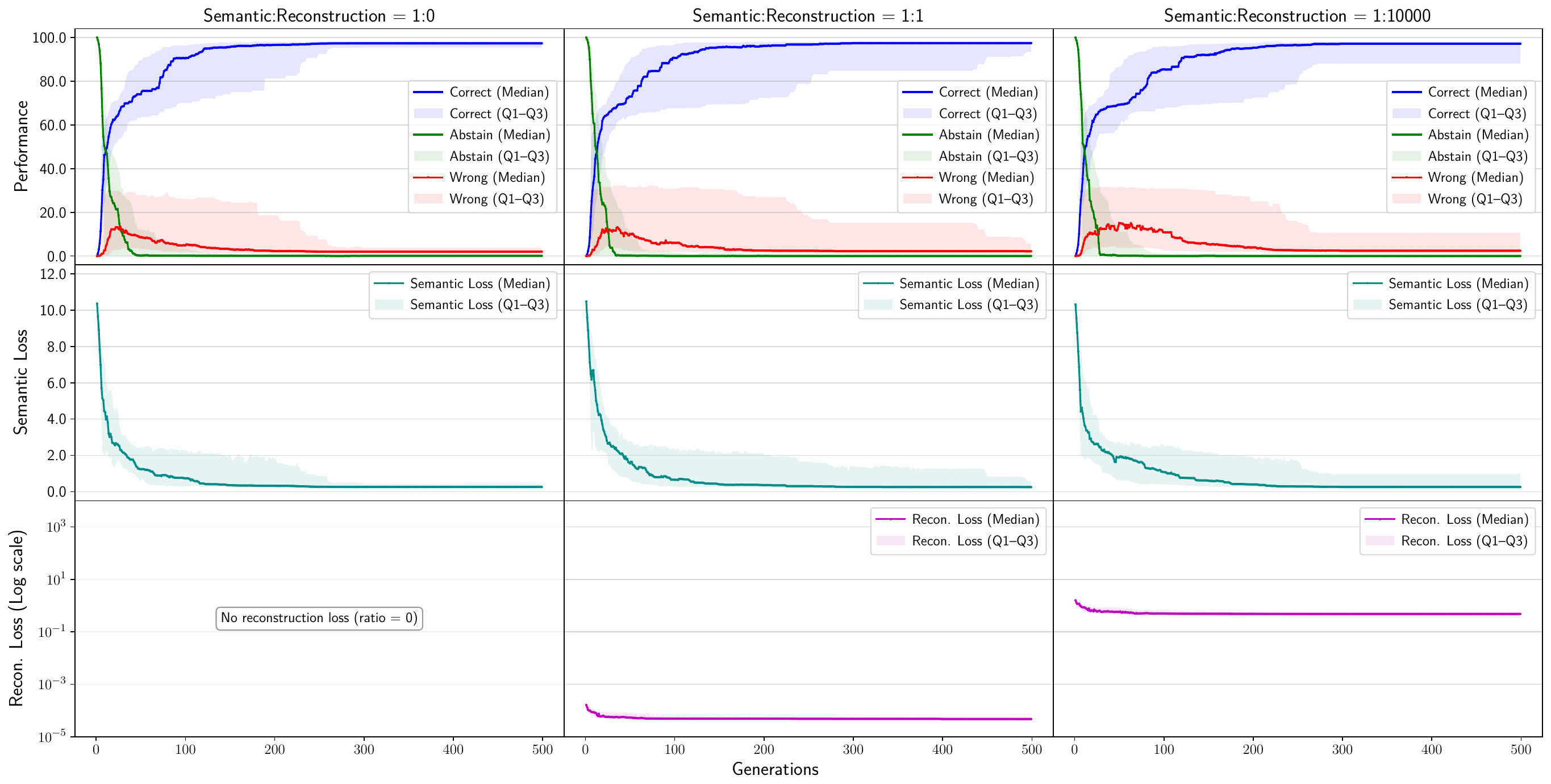}
    \caption{
        Aggregated results comparing three semantic:reconstruction loss ratios (1:0, 1:1, and 1:10000) on their respective \testSet subsets.
        Results are aggregated by generation number: for each generation, data from all experiments that reached that generation are combined.
        Since experiments terminate at different generations, later generations aggregate fewer experiments.
        The top subplot shows performance metrics (correct, abstain, and wrong predictions), while the middle and bottom subplots show semantic and reconstruction losses, respectively.
        Shaded areas depict the interquartile range (Q1--Q3) for each metric.
    }
    \label{fig:non-aggregated-results-compare-loss-ratios}
\end{figure*}

\Cref{fig:exp-1-normal} shows a typical example of an experimental run (further examples are included in Appendix~\ref{appendix:detailed-results}).
The upper subplot shows the performance against \validationSet and \testSet of the lineage of \fittest organisms selected by the evolutionary process.
As a typical example, the correct performance (blue line) of the organisms gradually increases, with a corresponding decrease in abstain performance (green line).
The wrong performance (red line) increases slightly early in the process, but eventually falls back to near zero.

\Cref{fig:aggregated-results} shows the aggregated results of all experiments on their respective testing subsets \testSet.
The top subplot shows performance metrics across training progress: correct performance (blue) rises steadily from near-zero to approaching $\sim$99\% median by the end of training, while abstain performance (green) follows an inverse trajectory, decreasing from initial dominance to near-zero as symbolic policies develop.
Wrong performance (red) remains relatively low throughout, with median values staying below 15\% and eventually declining as correct performance improves.
The middle and bottom subplots show the corresponding decreases in both semantic and reconstruction losses.
The reduction in reconstruction loss reflects neural learning, while the decrease in semantic loss indicates successful joint optimization of both components.

\subsection{Performance Optimization Validation}
\label{sec:perf-opt-validation}

To empirically validate the effectiveness of the technical contributions described in \Cref{sec:technical-contribs}, we conducted a series of experiments measuring their impact on computational performance.
The validation experiments trained 100 NeSy organisms across 20 independent runs (10 symbolic policies, each run twice) under various configurations of parallel processing and operation caching.
The total training times were recorded for each configuration, with results presented in \Cref{fig:parallel-gpu-training-operation-caching}.

Operation caching during semantic loss calculation provides substantial speedup across all parallelization configurations.
Comparing uncached (blue boxes) with cached (orange boxes) results in \Cref{fig:parallel-gpu-training-operation-caching} shows that caching the WMC computational graph consistently and significantly reduces training times.
This shows that repeated WMC calculations represent a major computational bottleneck that caching effectively addresses.

Parallel training of multiple neural networks on a single GPU yields additional performance gains.
The results show that increasing parallelization from sequential execution to approximately 5 parallel processes substantially decreases total experiment duration.
Beyond this point, efficiency gains plateau, likely due to GPU resource saturation and the overhead of managing concurrent processes.
The comparable performance between sequential training (``No multiproc.'') and single-process execution (1 process) confirms that the multiprocessing framework itself introduces minimal overhead.

These validation results demonstrate that both optimizations are essential for the feasibility of the large-scale evolutionary experiments presented in this paper.
For the main experiments reported in \Cref{sec:experimental-setup}, we adopted a configuration of 4 parallel processes per GPU, balancing computational efficiency with resource utilization based on the observed performance characteristics.

\subsection{Ablation Study: Loss Ratio Comparison} \label{sec:ablation-loss-ratios}

As motivated in \Cref{subsec:extended-neurolog}, the reconstruction loss was introduced to provide an additional training signal when abductive feedback is limited during early evolutionary generations.
To evaluate whether this auxiliary signal improves convergence, ablation experiments compared three different semantic:reconstruction loss ratios: 1:0 (semantic loss only), 1:1 (equal weighting), and 1:10000 (heavily weighted reconstruction loss).
Results are shown in \Cref{fig:non-aggregated-results-compare-loss-ratios}, which aggregates results by generation number (unlike \Cref{fig:aggregated-results}, which interpolates results to a unified scale).

The same general trend observed in individual experiments (e.g., \Cref{fig:exp-1-normal}) holds across all loss ratio configurations: during the evolutionary process, there is a gradual increase in correct performance with a simultaneous reduction in abstention performance.
Due to the greedy nature of the evolutionary algorithm, which always selects a single organism as the \fittest of each generation, there is an increase in wrong performance early in the evolutionary process, which later decreases to near zero as correct performance increases.

To assess whether different loss ratios affect evolutionary efficiency, we compared cumulative correct performance across all generations for each experiment run (giving one summary value per experiment), grouped by loss ratio.
This cumulative metric captures the overall evolutionary efficiency: configurations that reach high performance faster accumulate higher cumulative scores, even if final performance is similar.
Statistical analysis using Dunn's non-parametric test revealed no statistically significant differences across the three loss ratio configurations (all pairwise $p > 0.05$), as detailed in \Cref{tab:reconLossEpochs5}.
This indicates that the reconstruction loss component, despite being motivated by theoretical considerations about incomplete abductive feedback, does not provide measurable benefit for convergence.
The implications of this finding are discussed in \Cref{sec:discussion}.

\begin{table}[t]
    \centering
    \caption{
        Dunn's test p-values for cumulative correct performance comparisons between semantic:reconstruction loss ratios (1:0,~1:1,~1:10000).
        No significant differences were detected (all values~$> 0.05$).
    }
    \label{tab:reconLossEpochs5}
    \begin{tabular}{r|rrr}
        \toprule
        Ratio & 0    & 1    & 10000 \\
        \midrule
        0     & 1.00 & 0.78 & 0.87  \\
        1     & 0.78 & 1.00 & 0.91  \\
        10000 & 0.87 & 0.91 & 1.00  \\
        \bottomrule
    \end{tabular}
\end{table}

\subsection{Baselines} \label{sec:baselines}

Due to the specific nature of the policies used in this study, direct comparison with other NeSy frameworks in the literature was not feasible.
Frameworks such as MetaAbd~\citep{dai2021MetaAbd}, NSIL~\citep{cunnington2021InductiveLearningComplex}, and NeuralFastLAS~\citep{charalambous2023NeuralFastLASFastLogicBased} employ different policy structures, learning paradigms, and assumptions about knowledge representation that make them incompatible with our evolutionary approach to learning non-differentiable shallow propositional policies.
Instead, we compared our approach against an end-to-end neural baseline; specifically, the same CNN architecture used in the \neurMod of our NeSy organisms, but trained directly on the labeled data without any symbolic component.

\begin{figure}[t]
    \centering
    \includegraphics[width=1\linewidth]{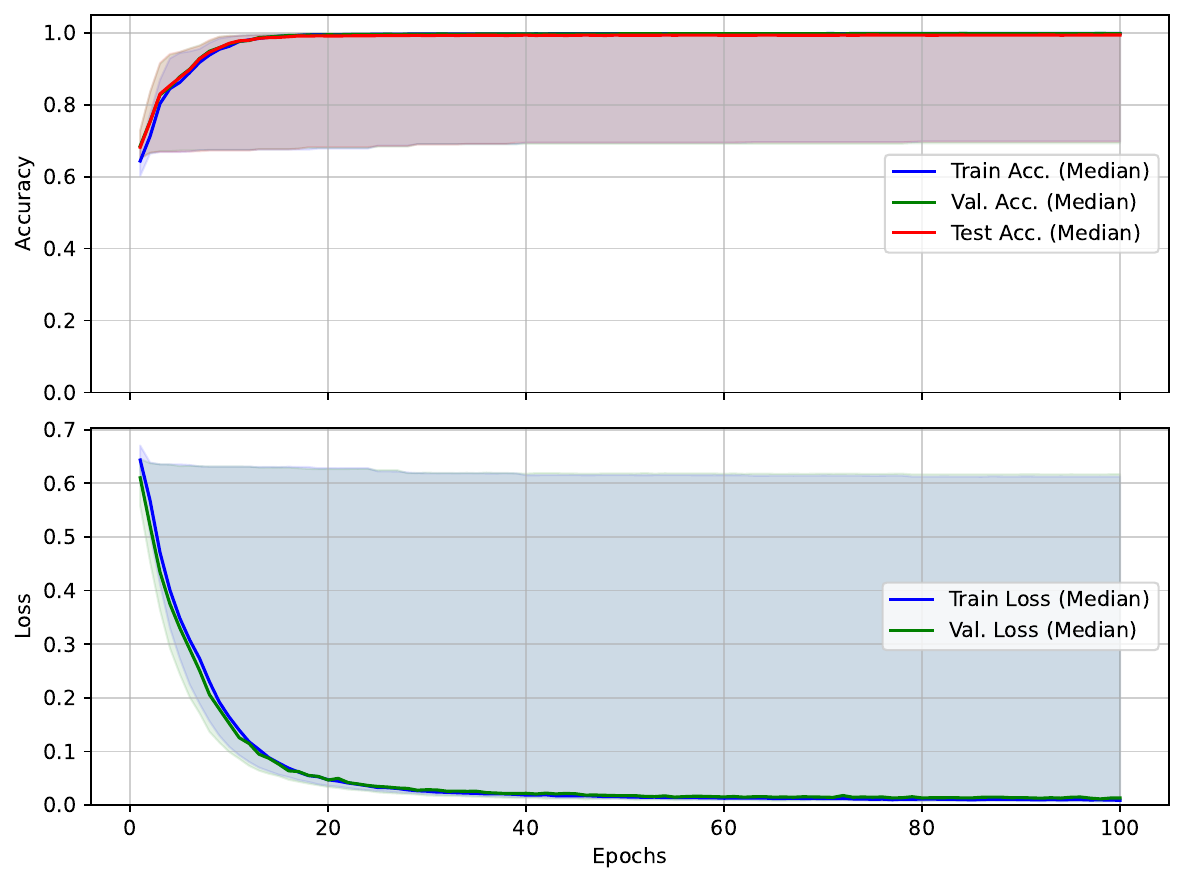}
    \caption{
        Training progression of the end-to-end neural baseline over 100 epochs across 150 datasets (30 target policies $\times$ 5 random dataset generations each).
        The baseline uses the same CNN architecture as the NeSy organisms' \neurMod but is trained directly on labeled data without symbolic components.
        Networks were trained using cross-entropy loss with the Adam optimizer.
        Performance metrics show accuracy on \trainingSet (blue), \validationSet (green), and \testSet (red), with median accuracy $>98\%$ on all sets.
        The lower subplot shows training and validation losses.
        Shaded areas depict the interquartile range (Q1--Q3) for each metric.
        The large variation in ranges shows that while most networks learn very well (median accuracy approaching 100\%), occasionally networks fail to converge effectively, resulting in lower accuracy and higher losses.
    }
    \label{fig:e2e-baseline-results}
\end{figure}

The end-to-end neural baseline was evaluated on the same 150 datasets used for the NeSy experiments (30 target policies $\times$ 5 random dataset generations each, as described in \Cref{sec:experimental-setup}), with identical training/validation/test splits.
The baseline was trained using the Adam optimizer with learning rate 0.001 and identical batch sizes to the NeSy organisms.
The baseline network receives sequences of 8 MNIST digit images as input and is trained to predict the binary label directly through supervised learning, without any intermediate symbolic reasoning.
While the base CNN encoder architecture remains identical to that used in the \neurMod (processing each of the 8 images sequentially to produce n-dimensional output vectors), the end-to-end baseline includes additional layers for direct binary classification: the concatenated encoder outputs are flattened (producing an 8n-dimensional vector) and passed through two fully-connected layers (64 neurons with ReLU activation, then 2 output neurons) followed by softmax, enabling direct prediction of \specialAtom or \neg{\specialAtom} without symbolic reasoning.
\Cref{fig:e2e-baseline-results} shows the progression of the training of the end-to-end neural baseline over 100 epochs.
Although median performance reaches $>98\%$ accuracy on all sets, the results exhibit substantial variance between networks, with occasional convergence failures contributing to the large interquartile ranges visible in the figure.

\section{Discussion} \label{sec:discussion}

\begin{figure*}[htb]
    \centering
    \includegraphics[width=\linewidth]{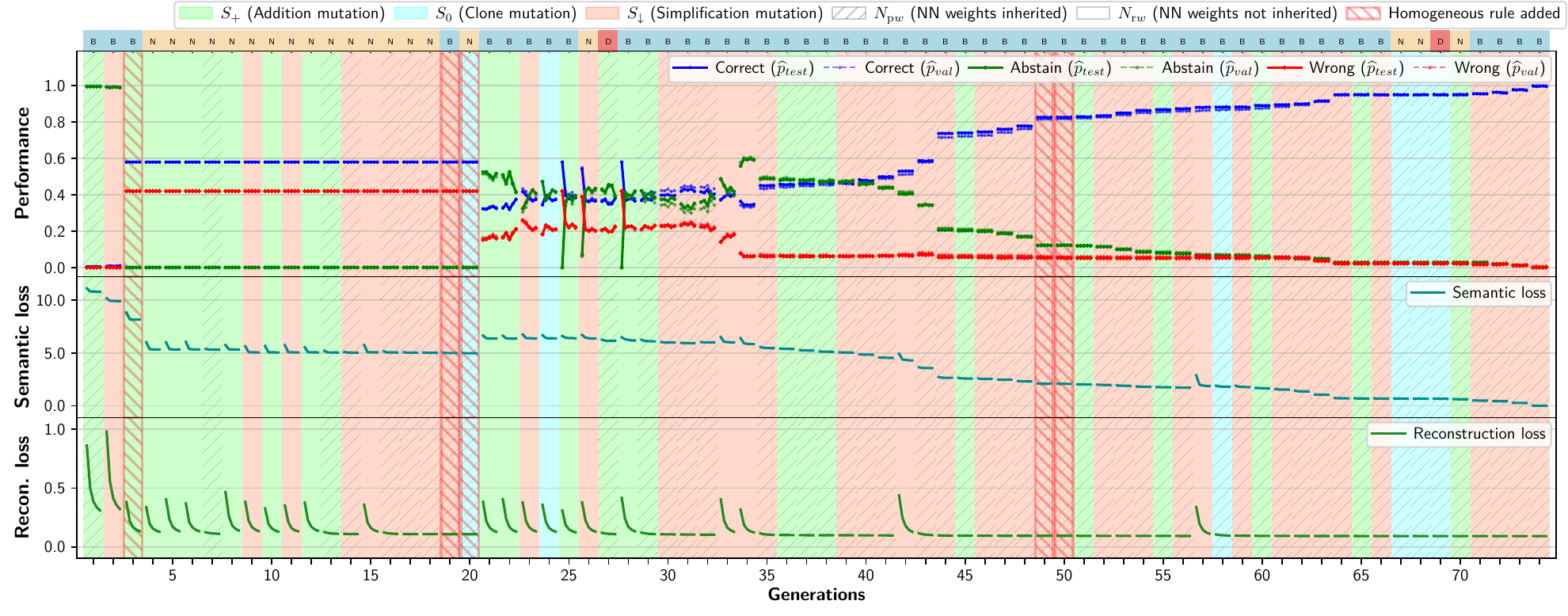}
    \caption{
        Experimental run showing the stuck state for part of the evolutionary process.
    }
    \label{fig:exp-2-stuck}
\end{figure*}

The results presented in \Cref{sec:main-results} demonstrate that the proposed evolutionary framework successfully enables NeSy systems to learn non-differentiable symbolic policies while concurrently training neural networks.
Starting from empty symbolic policies and randomly initialized neural weights, the evolutionary process consistently discovers policies that approximate hidden targets, with a median correct performance approaching 99\% by the end of training (\Cref{fig:aggregated-results}).
The characteristic pattern observed across experiments---correct performance rising steadily while abstentions decrease (\Cref{fig:exp-1-normal})---indicates that the evolutionary process progressively builds symbolic policies that cover an increasing portion of the input space.
The simultaneous decrease in both losses confirms effective learning.
Reconstruction loss decrease indicates neural learning through self-supervision, while semantic loss decrease reflects successful joint optimization, as this measure depends on both neural network outputs and the symbolic policy's abductive feedback.

These findings validate the central hypothesis of this work: that non-differentiable symbolic policies can be learned from scratch through evolutionary search, without requiring predefined symbolic knowledge or gradient-based policy optimization.
The fact that the majority of experiments converge to near-perfect correct performance demonstrates the viability of this approach for concurrent neural-symbolic learning.

Although the majority of the experiments followed this successful convergence pattern, in some cases the evolutionary process entered what we term a \emph{stuck state}, i.e.\ a local optimum where organisms achieve partial correct performance on training and validation data, but fail to improve further across subsequent generations.
\Cref{fig:exp-2-stuck} shows such an example: the evolutionary process falls into a stuck state at generation 3.
The stuck state persists until generation 21, whereupon it is broken, allowing the evolutionary process to sequentially select a lineage of organisms that reach a correct performance of \(>99\%\) at generation 74.

Stuck states result from the addition of a \notion{homogeneous rule}---a rule whose body contains literals that all share the same sign---to an organism's policy via \mutationAdd or \mutationSimplify mutations.
Examples of homogeneous rules include ``\atom{a1}, \atom{a2} \impliesRule \atom{head}'' (all positive atoms) or ``\atom{-a1}, \atom{-a2} \impliesRule \atom{head}'' (all negative atoms).
When such rules are added to a policy, the \neurMod sometimes converges to a trivial solution where it classifies all input images into the same category (all as digit~1 or all as digit~2), regardless of the actual input content.
This uniform classification causes the homogeneous rule to fire consistently, leading the organism to produce the same prediction (\specialAtom or \neg{\specialAtom}) for all data instances.
Consequently, the organism achieves misleadingly elevated relative fitness by ``correctly'' predicting the proportion of \trainingSet and \validationSet labeled with the same atom (\atom{head} or \atom{-head}) that its \neurMod uniformly predicts.
This high fitness leads to preferential selection of such organisms, trapping the evolutionary process in a local optimum (a more detailed technical explanation is given in Appendix~\ref{appendix:detailed-results}).

The stuck state phenomenon, combined with the challenge of training neural networks when symbolic policies are incomplete during early evolutionary generations, motivated the introduction of reconstruction loss as a self-supervised learning mechanism.
As mentioned in \Cref{subsec:extended-neurolog}, when organisms possess minimal symbolic policies, the abductive feedback available for calculating semantic loss is limited, potentially hindering effective neural learning.
The framework incorporated a reconstruction loss component with Gumbel-Softmax regularization to provide an additional training signal during these challenging phases.
However, the results of the ablation study reported in \Cref{sec:ablation-loss-ratios} demonstrate that the semantic loss alone is sufficient for successful convergence, indicating that abductive feedback from the symbolic module, even when policies are initially incomplete, provides an adequate training signal for the neural networks.
This simplifies the proposed framework: the reconstruction loss component, despite being initially incorporated on the basis of theoretical motivations, is not necessary for achieving strong performance.
The framework can operate effectively with semantic loss alone, reducing architectural complexity and eliminating the computational overhead associated with the decoder network and reconstruction loss calculation.

Ultimately, given enough generations, the majority of stuck experiments eventually break out of the stuck state.
This is evident in \Cref{fig:aggregated-results}, where at the end of training, the \textit{median} correct performance approaches \(100\%\), while the corresponding \textit{mean} performance trails slightly behind (while still following an upward trend), indicating a very small percentage of stuck experiments at the end of training.
The interquartile range (Q1--Q3) shown in the shaded regions of \Cref{fig:aggregated-results} shows variance across experiments, which arises from the stochastic nature of both the evolutionary process (random rule generation and selection) and the neural network training (weight initialization and batch sampling).
This variance reflects the inherent exploration-exploitation trade-off in evolutionary search: some lineages discover effective symbolic policies more quickly through fortuitous mutations, while others require more generations to escape suboptimal regions of the search space.

Given the strong performance of the end-to-end neural baseline described in \Cref{sec:baselines}, a natural question arises: \emph{Why pursue the evolutionary NeSy approach when the end-to-end neural baseline achieves comparable or superior performance with substantially lower computational cost?}
The baseline achieves a median accuracy exceeding 98\% with a single network trained for 100 epochs, while the evolutionary framework requires training multiple organisms across many generations, representing a significant difference in computational expense.
This disparity reflects the computational \emph{price of interpretability} mentioned in the Introduction, understood as the additional computational resources and longer training times required to obtain human-readable symbolic policies alongside neural components.

The value proposition of this trade-off lies in the interpretable symbolic policies that emerge from the evolutionary process.
Unlike the black-box nature of pure neural networks, the learned symbolic policies provide explicit, inspectable logical rules that explain the system's reasoning.
These policies can be examined by domain experts without machine learning expertise, modified if necessary, and verified for consistency with domain knowledge.
The symbolic component offers transparency that is fundamentally absent in end-to-end neural approaches, regardless of their predictive performance.

This trade-off becomes justifiable in domains where interpretability and trust are critical.
In safety-critical applications (medical diagnosis, autonomous systems), legal and regulatory contexts (credit decisions, criminal justice), and scientific domains that require explainable models, the ability to inspect and validate reasoning processes may outweigh computational efficiency concerns.
When decisions must be explained to stakeholders, audited for compliance, or validated by domain experts, symbolic policies provide a level of transparency that end-to-end neural baselines cannot match.
The additional computational cost provides verifiable symbolic policies that enable accountability and validation by domain experts, which may be essential in such contexts.

\section{Related Work} \label{sec:relatedWork}

Early approaches to neural-symbolic integration focused on making symbolic reasoning differentiable to enable gradient-based learning.
\citet{serafini2016LogicTensorNets} introduced Logic Tensor Networks (LTNs), mapping logical constructs to differentiable tensor operations, while \citet{evans2018LearningExplanatoryRules} proposed $\partial$ILP, embedding symbolic rules as differentiable components amenable to backpropagation through forward-chaining deduction implemented as differentiable computation.
These foundational works demonstrated that symbolic reasoning could be integrated with deep learning through differentiability, although both required the symbolic component to support gradient propagation.

More recent work has advanced joint learning of neural perception and symbolic reasoning.
\citet{mao2019NSCL} introduced the Neuro-Symbolic Concept Learner (NS-CL), demonstrating how disentangled perception and symbolic reasoning yield transparent, compositional visual understanding through joint acquisition of visual concepts, word meanings, and semantic parsing.
\citet{dai2021MetaAbd} proposed MetaAbd, which jointly learns subsymbolic perception and symbolic reasoning from raw data within an expectation-maximization framework, successfully inducing reusable logic programs with advantages in predictive accuracy and data efficiency over end-to-end neural baselines.

Building on this paradigm, \citet{cunnington2021InductiveLearningComplex,cunnington2024NeuroSymbolicLearningAnswer} developed NSIL, which alternates between neural learning of latent concepts and Answer Set Programming hypothesis learning, later extending it to handle unlabeled raw data.
\citet{charalambous2023NeuralFastLASFastLogicBased} introduced NeuralFastLAS, training neural networks alongside learned rule posteriors with semantic loss, achieving high accuracies with reduced training times.
\citet{daniele2023DSL} proposed Deep Symbolic Learning (DSL), which departs from relying on given symbolic knowledge by jointly learning perception and symbol composition functions within a differentiable pipeline, making discrete symbolic choices through policy functions inspired by reinforcement learning.

Complementary advances in automatic predicate invention have emerged from \citet{sha2023NeSyPredicateInvention}, who introduced NeSy-\(\pi\) for automatic predicate invention from visual scenes, and \citet{barbiero2023DCR}, whose Deep Concept Reasoner generates and evaluates fuzzy logic rules from concept embeddings without explicit concept supervision, demonstrating semantically consistent reasoning learned directly from embeddings.

Integration approaches have diversified across methodologies.
\citet{diaz-rodriguez2022EXplainable} fused deep learning with domain knowledge graphs through XAI-informed training aligning feature attributions with knowledge.
\citet{pryor2023NeuPSL} introduced NeuPSL, extending probabilistic soft logic with neural predicates for end-to-end gradient training.
\citet{zhong2023ChatABLAbductiveLearning} integrated large language models into abductive learning through ChatABL, transforming visual features into natural language logical facts for LLM-based reasoning.

Recent surveys have provided systematic perspectives on the field.
\citet{vermeulen2023Experimental} categorized NeSy tasks and demonstrated that probabilistic logic-programming approaches achieve superior performance at higher computational cost, while \citet{marra2024Statistical} proposed the NeSy recipe, a unified two-phase pipeline facilitating systematic comparison and design across approaches.

While these approaches have advanced neural-symbolic integration, most assume predefined symbolic structures, rule templates, or background knowledge.
Our work addresses this limitation through evolutionary learning, starting with empty policies and progressively building symbolic knowledge through mutation and selection without requiring differentiability or predefined structures.
This evolutionary approach to concurrent neural and symbolic learning offers a complementary direction for NeSy systems, particularly in domains where symbolic knowledge acquisition from experts is challenging.

\section{Conclusions \& Future Work} \label{sec:conclusionsFutureWork}

In this study, we propose a new framework that facilitates the simultaneous training of both neural and symbolic components within NeSy systems.
Our extensive experimentation validates the effectiveness of this approach, demonstrating that NeSy systems can start with empty symbolic policies and randomly initialized neural weights, and progressively evolve to closely approximate hidden target policies.
Arguably, the most important contribution of our work is that it demonstrates that it is feasible to learn non-differentiable policies, while simultaneously training neural networks in NeSy systems.
The demonstrated capability for evolvable policies within NeSy systems may be a first step towards facilitating research in areas where symbolic knowledge acquisition from domain experts is challenging.

Although our ablation study on reconstruction loss demonstrated its dispensability, future work could explore the individual contributions of other framework components to performance metrics.
Additionally, it will be explored whether evolution can be replaced by other search methods, for example through the use of less greedy versions of the evolutionary process that allow the retention of different forms diversity in the population~\citep[e.g., quality diversity optimization techniques, see][]{chatzilygeroudis2021QualityDiversity,mouret2015MappingElites,vassiliades2018VoronoiPhenotypic}.
Another research avenue would be to experiment with larger target policies (in terms of number of binary concepts), and the use of the proposed framework with data from real-world scenarios.

\begin{funding}
    This project has received funding from the European Union's Horizon 2020 Research and Innovation Programme, under Grant agreement No 739578, complemented by the Government of the Republic of Cyprus through the Directorate General for European Programmes, Coordination and Development.
\end{funding}

\begin{acks}
    We are grateful to Kyriacos Mosphilis and Vasileios Markos for their insightful discussions throughout the project.
\end{acks}

\begin{sm}
    The source code, experimental configurations, and data generation scripts for reproducing all experiments presented in this paper are publicly available at \githubRepo.
\end{sm}

\theendnotes

\bibliographystyle{SageH}
\bibliography{bibliography.bib}

\pagebreak

\appendix

\section{Empirical Setup (Details)}\label{appendix:empirical-setup-detailed}

The relative fitness during the evolutionary process was calculated using the score matrix shown in \Cref{tab:rel_fitness_matrix}.

\begin{figure*}
    \centering
    \includegraphics[width=0.75\linewidth]{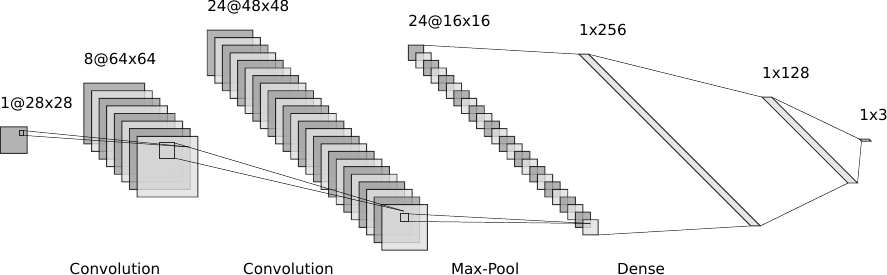}
    \caption{
        CNN architecture used in the experiments.
        Diagram created using NN-SVG~\citep{lenail2019NNSVG}.
    }
    \label{fig:nn-arch}
\end{figure*}

The CNN architecture used in the experiments is shown in \Cref{fig:nn-arch}.
This base encoder architecture was used in both the NeSy organisms' \neurMod and the end-to-end baseline.
Since each data instance was made up of eight MNIST images, the same NN was used eight times sequentially to obtain predictions for all input images.
For the NeSy organisms, the encoder outputs n neurons per image (8 atoms), producing 8n outputs total that are concatenated and used as symbolic atom activations.
For the end-to-end baseline, these 8n concatenated outputs are instead flattened and passed through two additional fully-connected layers: a hidden layer with 64 neurons and ReLU activation, followed by an output layer with 2 neurons (representing \specialAtom and \neg{\specialAtom}), with softmax applied for binary classification.

Each experiment was run in a Rocky Linux 8.5 environment, using Python v3.11.6, with 4 CPU cores and 48GB RAM, and an Nvidia RTX A5000 GPU with 24GB VRAM.

\begin{table}[ht]
    \caption{
        Score matrix used for the calculation of relative fitness.
    }
    \label{tab:rel_fitness_matrix}
    \centering
    \input{assets/tables/relative_fitness}
\end{table}

\section{Results (Details)} \label{appendix:detailed-results}

\subsection{Aggregated Results}

Detailed aggregated results at the final generation for all experiments are shown in \Cref{tab:resultsTable}.

\begin{table}[htb]
    \centering
    \caption{
        Aggregated results at the end of all experiments on \testSet.
    }
    \label{tab:resultsTable}
    \begin{tabular}{l|rrr}
        \toprule
        {}      & Median & Mean  & SD    \\
        \midrule
        Correct & 0.992  & 0.944 & 0.109 \\
        Abstain & 0.000  & 0.002 & 0.003 \\
        Wrong   & 0.005  & 0.053 & 0.110 \\
        \bottomrule
    \end{tabular}
\end{table}

\subsection{Stuck State}

\Crefrange{fig:exp-stuck-1}{fig:exp-stuck-4} show the results of individual experimental runs that exhibit the stuck state for part of their evolutionary process; in all these cases the stuck state was eventually broken.

During some evolutionary runs, organisms entered \emph{stuck states} (local optima where they achieve partial correct performance on training and validation data, but fail to improve further across subsequent generations).
Organisms become stuck after a symbolic \mutationAdd or \mutationSimplify mutation adds a \emph{homogeneous rule} to their policy.
A homogeneous rule is a rule whose body contains literals that all share the same sign, such as ``\atom{a1}, \atom{a2}, \atom{a3} \impliesRule \atom{head}'' (all positive atoms) or ``\atom{-a1}, \atom{-a2}, \atom{-a3} \impliesRule \atom{head}'' (all negative atoms).

The addition of such rules triggers an adverse interaction between the homogeneous symbolic structure and the sequential neural architecture.
Specifically, the \neurMod converges to a trivial solution where it classifies all input images into the same category (all as digit~1 or all as digit~2), regardless of the actual input content.
This uniform classification causes the homogeneous rule in the symbolic policy to fire consistently for all data instances, causing the \symMod to always produce the same prediction (\specialAtom or \neg{\specialAtom}).
Consequently, the organism achieves misleadingly high relative fitness by ``correctly'' predicting the proportion of \trainingSet and \validationSet labeled with the atom (\atom{head} or \atom{-head}) that its \neurMod uniformly outputs.

This high fitness leads to preferential selection of such organisms during the evolutionary process.
The stuck state persists because it is not often that subsequent mutations produce offspring that exceed the local optimum fitness achieved by organisms with homogeneous rules.

The pattern described above is due to an adverse interaction between the NN architecture used in the \neurMod, and incomplete hypothesis policies that give one-sided signals to it during training, due to the addition of homogeneous rules to the hypothesis policies.
Since each image of each atom is processed by the NN sequentially (see \Cref{fig:neurolog-deduction}), the signals given by the homogeneous rule sometimes lead the NN to always predict the same value regardless of input.

However, the occurrence of the pattern is limited to \(\sim\)\(15\%\) of the experiments at any given time, and the percentage decreases as the evolutionary process continues, indicating that the pattern is eventually broken in the majority of the experiments.
This is evident in \Cref{fig:aggregated-results}, where at the end of training the \textit{median} correct performance approaches \(100\%\), while the corresponding mean performance trails slightly behind (while still following an upward trend), indicating a certain number of experiments that conclude while still stuck.
Empirically, it was observed that most instances of stuck experiments eventually break out of the stuck state, given enough generations.
\Cref{fig:exp-2-stuck} shows an example of such an experimental run, that becomes stuck early in the evolutionary process, but eventually becomes unstuck, and finally reaches almost \(100\%\) correct performance.

\subsection{Individual Experiment Plots}

\begin{figure*}[htb]
    \centering
    \includegraphics[width=\linewidth]{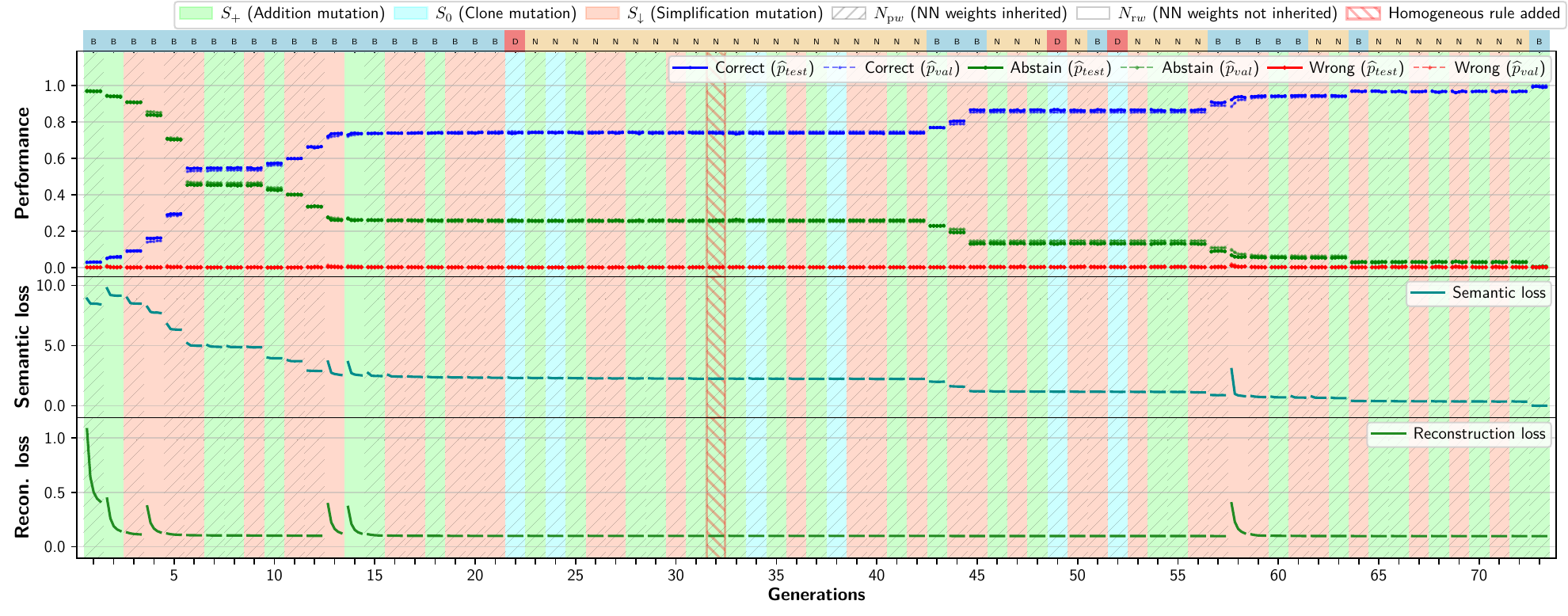}
    \caption{Typical experimental run.}
    \label{fig:exp-normal-1}
\end{figure*}

\begin{figure*}[htb]
    \centering
    \includegraphics[width=\linewidth]{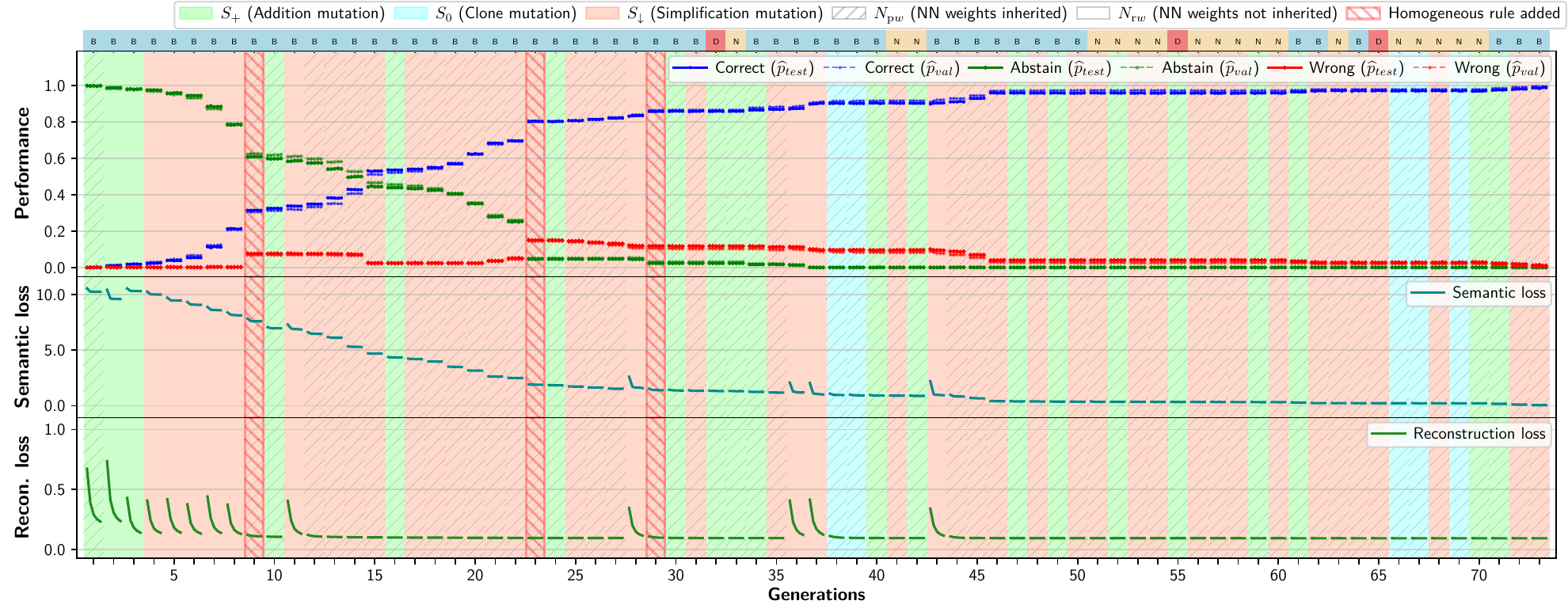}
    \caption{Typical experimental run.}
    \label{fig:exp-normal-2}
\end{figure*}

\begin{figure*}[htb]
    \centering
    \includegraphics[width=\linewidth]{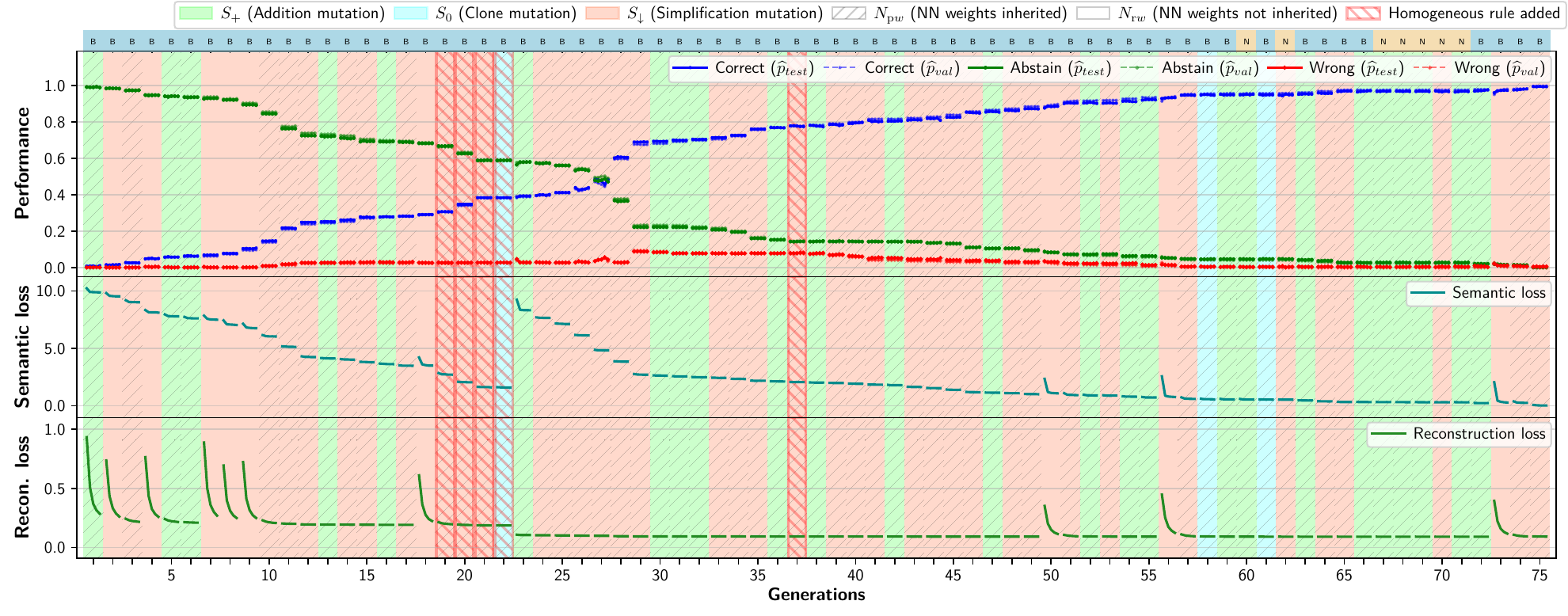}
    \caption{Typical experimental run.}
    \label{fig:exp-normal-3}
\end{figure*}

\begin{figure*}[htb]
    \centering
    \includegraphics[width=\linewidth]{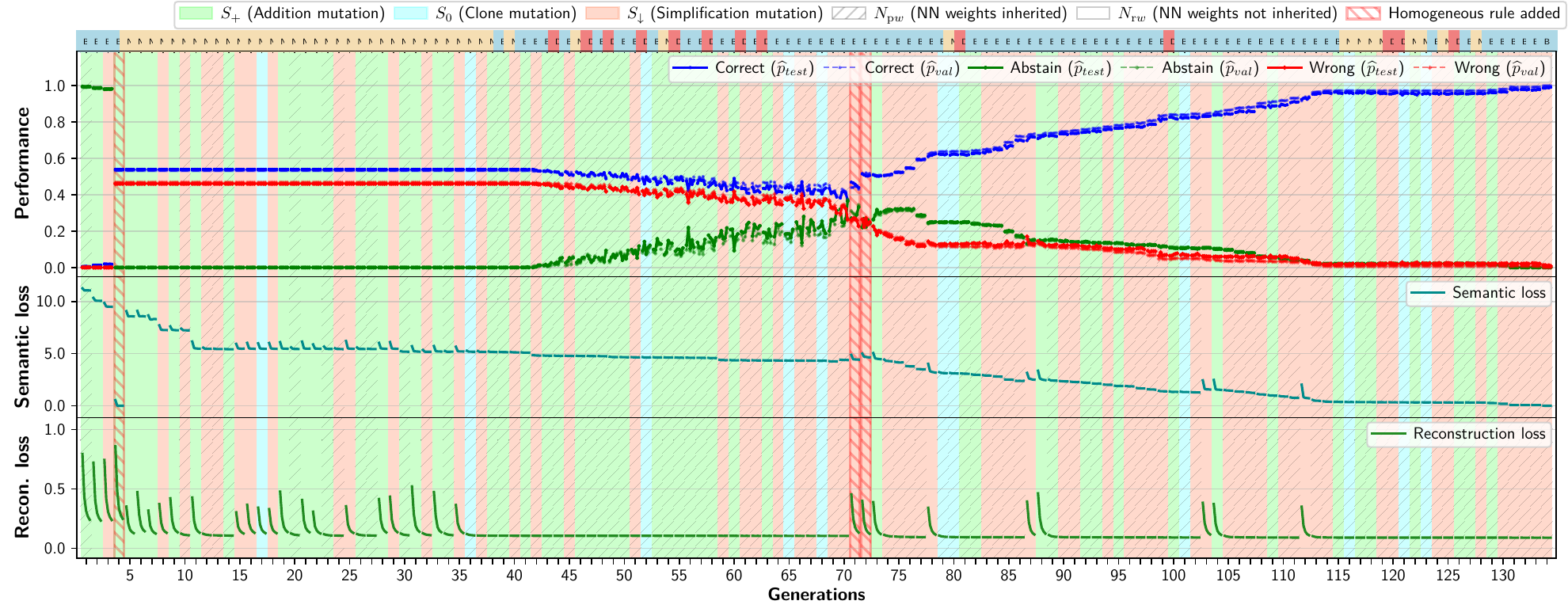}
    \caption{Experimental run showing the stuck state for part of the evolutionary process.}
    \label{fig:exp-stuck-1}
\end{figure*}

\begin{figure*}[htb]
    \centering
    \includegraphics[width=\linewidth]{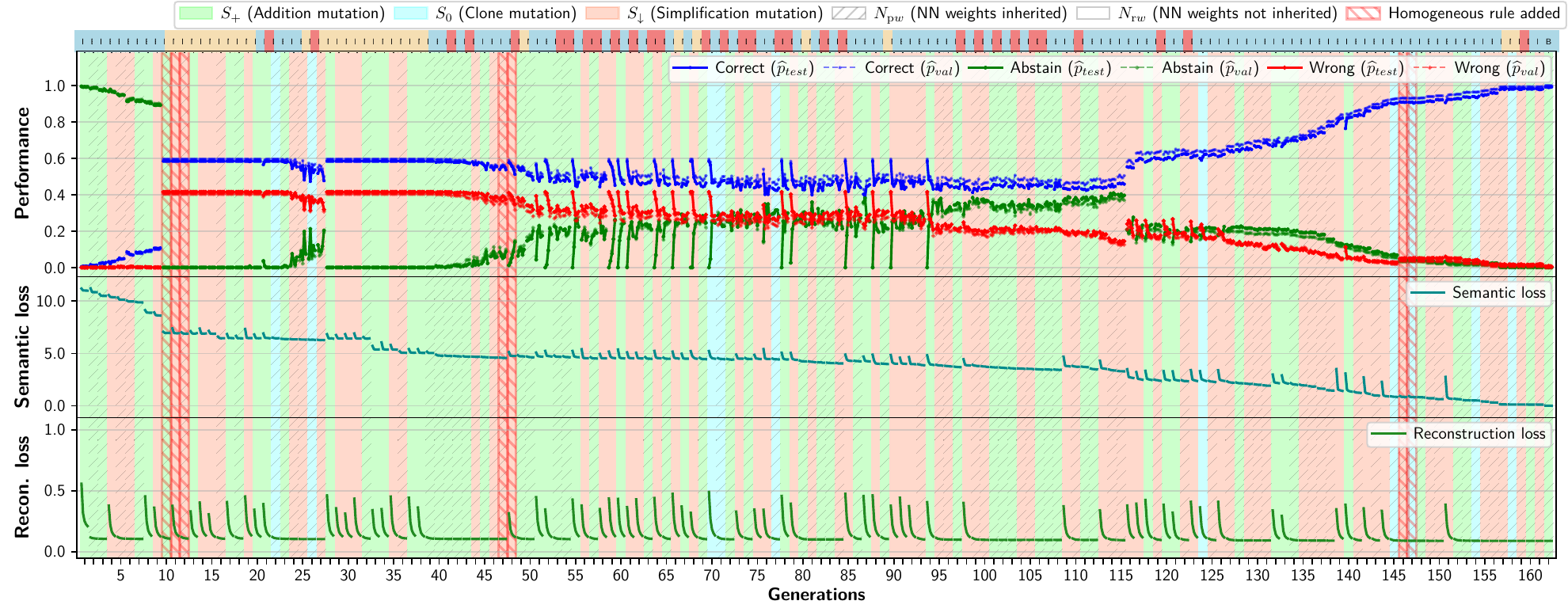}
    \caption{Experimental run showing the stuck state for part of the evolutionary process.}
    \label{fig:exp-stuck-2}
\end{figure*}

\begin{figure*}[htb]
    \centering
    \includegraphics[width=\linewidth]{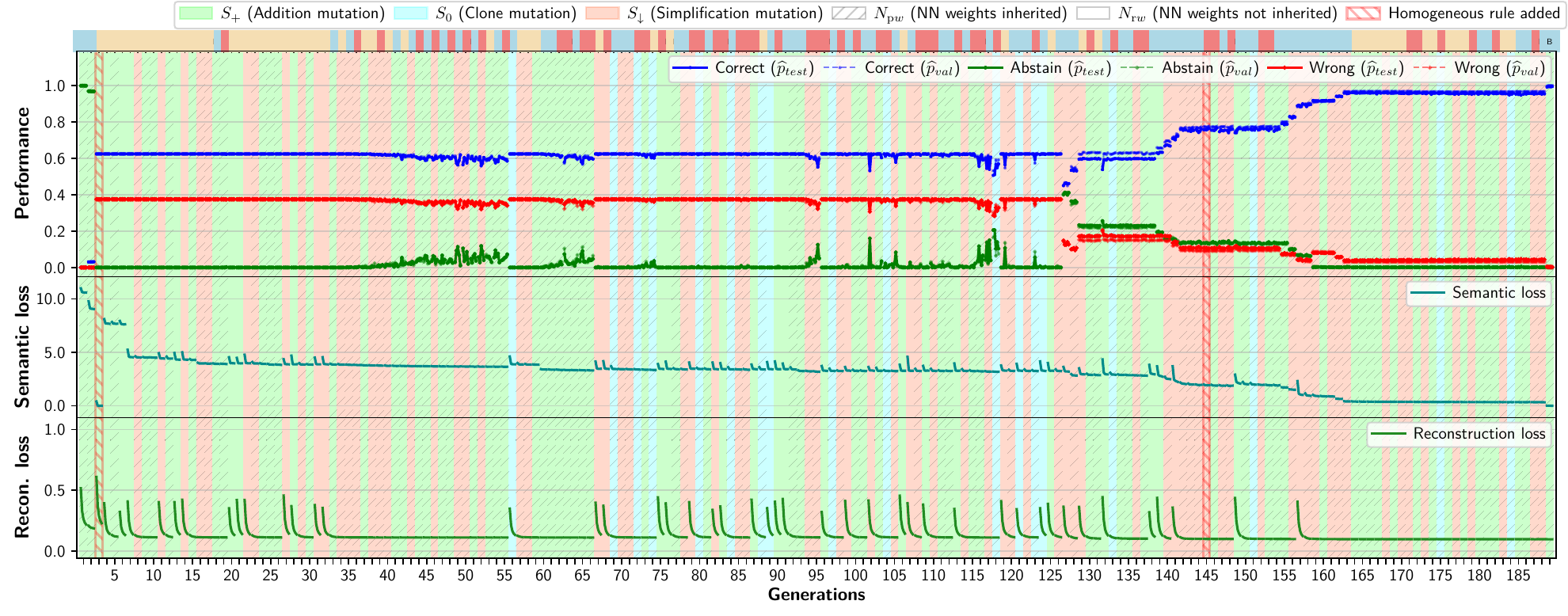}
    \caption{Experimental run showing the stuck state for part of the evolutionary process.}
    \label{fig:exp-stuck-3}
\end{figure*}

\begin{figure*}[htb]
    \centering
    \includegraphics[width=\linewidth]{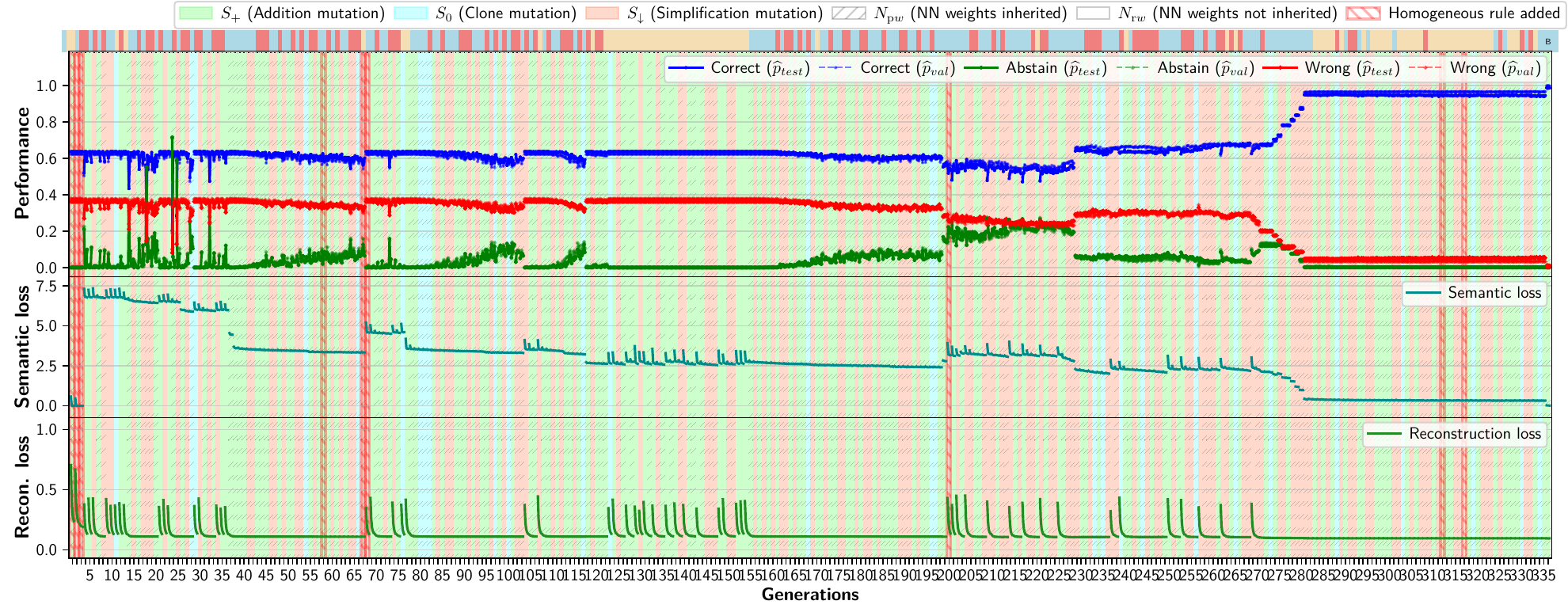}
    \caption{Experimental run showing the stuck state for part of the evolutionary process.}
    \label{fig:exp-stuck-4}
\end{figure*}

\Crefrange{fig:exp-normal-1}{fig:exp-normal-3} show the results of typical experimental runs, while \Crefrange{fig:exp-stuck-1}{fig:exp-stuck-4} show the results of individual experimental runs that exhibit the stuck state for part of their evolutionary process.


\end{document}

%% file: assets/figures/neurologTikz.tex
\tikzset{
    module/.style={rectangle, thick, draw=black, fill=black!1, align=center, font=\ttfamily, inner sep=4pt},
    module2/.style={rectangle, thick, draw=black, fill=black!1, align=flush left, font=\ttfamily, inner sep=0pt, minimum height=20mm, minimum width=40mm},
    method/.style={rectangle, align=center, draw=black, fill=white, font=\ttfamily},
    io/.style={rectangle, align=center},
    component/.style={rectangle, align=center, fill=red!5, draw=red, rounded corners=1ex},
    arrowTest/.style={-Stealth[], densely dashed, draw=blue},
    arrowTrain/.style={-Stealth[], dash dot, draw=red},
    arrowTrainLabel/.style={circle, solid, draw=red, font=\tiny\bfseries, text=white, fill=red!80, inner sep=0.5pt, outer sep=3pt},
    arrowTestLabel/.style={circle, solid, draw=blue, font=\tiny\bfseries, text=white, fill=blue!80, inner sep=0.5pt, outer sep=3pt}
}

\begin{tikzpicture}[
        auto,
    ]    
        \node[module] (nm) at (0,0) {Neural\\Module};
        \node[module] (sm) at (6,0) {Symbolic\\Module};
        \node[module] (trans) at (3,0) {Translator};
        \node[io] (input) at (-1.6,0) {Input};
        \node[io] (outputLabel) at (6+2.5,0) {Output/Label};
        \node[component] (semLoss) at (1.6,-1) {Loss calculation};
        
        
        \draw[arrowTest] (input.north) to [bend left] ([yshift=1mm]nm.west);
        \draw[arrowTest] ([yshift=1mm]nm.east) to [bend left] (trans.west);
        \draw[arrowTest] ([yshift=1mm]trans.east) to [bend left] ([yshift=1mm]sm.west);        
        \draw[arrowTest] ([yshift=1mm]sm.east) to [bend left]  (outputLabel.north);
        
        \draw[arrowTrain] (input.south) to [bend right] ([yshift=-1mm]nm.west);
        \draw[arrowTrain] (outputLabel.south) to [bend left] ([yshift=-1mm]sm.east);
        \draw[arrowTrain] ([yshift=-1mm]sm.west) to [bend left]  ([yshift=-1mm]trans.east);
        \draw[arrowTrain] ([yshift=-2mm]nm.east) to [bend left] (semLoss.north);
        \draw[arrowTrain] ([yshift=-1mm]trans.west) to [bend right] (semLoss.north);
        \draw[arrowTrain] (semLoss.west) to [bend left] (nm.south);
        
        
        \begin{scope}[on background layer]
            \node[draw=black!15, fill=black!5, 
                rounded corners=2ex, inner ysep=25pt, 
                inner xsep=8pt, anchor=north,
                fit=(nm) (trans) (sm)] (bg) {};
            \node[below] at (bg.north) {\textbf{\textsc{NeuroLog}}};
        \end{scope}
    \end{tikzpicture}

%% file: assets/tables/relative_fitness.tex
    {\renewcommand{\arraystretch}{1.2}%
        \begin{tabular}{cc|c|c|c|}
          & \multicolumn{1}{c}{} & \multicolumn{3}{c}{\textbf{Offspring}} \\
          & \multicolumn{1}{c}{} & \multicolumn{1}{c}{$Correct$} & \multicolumn{1}{c}{$Abstain$} & \multicolumn{1}{c}{$Wrong$} \\\cline{3-5}
          \parbox[t]{2mm}{\multirow{3}{*}{\rotatebox[origin=c]{90}{\textbf{Parent}}}} & $Correct$ & $0$  & $-1$ & $-1$ \\\cline{3-5}
                & $Abstain$ & $1$  & $0$  & $-1$ \\\cline{3-5}
                & $Wrong$   & $1$  & $1$  & $0$ \\\cline{3-5}
        \end{tabular}
    }
    

%% file: bibliography.bib
@inproceedings{barbiero2023DCR,
  title     = {Interpretable {{Neural-Symbolic Concept Reasoning}}},
  booktitle = {Proceedings of the 40th {{International Conference}} on {{Machine Learning}}},
  author    = {Barbiero, Pietro and Ciravegna, Gabriele and Giannini, Francesco and Zarlenga, Mateo Espinosa and Magister, Lucie Charlotte and Tonda, Alberto and Lio, Pietro and Precioso, Frederic and Jamnik, Mateja and Marra, Giuseppe},
  year      = {2023},
  month     = jul,
  pages     = {1801--1825},
  publisher = {PMLR},
  issn      = {2640-3498},
  urldate   = {2024-06-20}
}

@misc{bertsimas2019PriceInterpretability,
  title         = {The {{Price}} of {{Interpretability}}},
  author        = {Bertsimas, Dimitris and Delarue, Arthur and Jaillet, Patrick and Martin, Sebastien},
  year          = {2019},
  month         = jul,
  number        = {arXiv:1907.03419},
  eprint        = {1907.03419},
  primaryclass  = {cs},
  publisher     = {arXiv},
  doi           = {10.48550/arXiv.1907.03419},
  urldate       = {2025-11-07},
  archiveprefix = {arXiv}
}

@incollection{besold2021Chapter1,
  title      = {Chapter 1. {{Neural-Symbolic Learning}} and {{Reasoning}}: {{A Survey}} and {{Interpretation}}},
  shorttitle = {Chapter 1. {{Neural-Symbolic Learning}} and {{Reasoning}}},
  booktitle  = {Frontiers in {{Artificial Intelligence}} and {{Applications}}},
  author     = {Besold, Tarek R. and {d'Avila Garcez}, Artur and Bader, Sebastian and Bowman, Howard and Domingos, Pedro and Hitzler, Pascal and K{\"u}hnberger, Kai-Uwe and Lamb, Luis C. and Lima, Priscila Machado Vieira and De Penning, Leo and Pinkas, Gadi and Poon, Hoifung and Zaverucha, Gerson},
  editor     = {Hitzler, Pascal and Sarker, Md Kamruzzaman},
  year       = {2021},
  month      = dec,
  pages      = {1--51},
  publisher  = {IOS Press},
  doi        = {10.3233/FAIA210348},
  urldate    = {2023-11-17},
  isbn       = {978-1-64368-244-0 978-1-64368-245-7}
}

@misc{charalambous2023NeuralFastLASFastLogicBased,
  title         = {{{NeuralFastLAS}}: {{Fast Logic-Based Learning}} from {{Raw Data}}},
  shorttitle    = {{{NeuralFastLAS}}},
  author        = {Charalambous, Theo and Aspis, Yaniv and Russo, Alessandra},
  year          = {2023},
  month         = oct,
  number        = {arXiv:2310.05145},
  eprint        = {2310.05145},
  primaryclass  = {cs},
  publisher     = {arXiv},
  doi           = {10.48550/arXiv.2310.05145},
  urldate       = {2024-06-26},
  archiveprefix = {arXiv}
}

@incollection{chatzilygeroudis2021QualityDiversity,
  title      = {Quality-{{Diversity Optimization}}: {{A Novel Branch}} of {{Stochastic Optimization}}},
  shorttitle = {Quality-{{Diversity Optimization}}},
  booktitle  = {Black {{Box Optimization}}, {{Machine Learning}}, and {{No-Free Lunch Theorems}}},
  author     = {Chatzilygeroudis, Konstantinos and Cully, Antoine and Vassiliades, Vassilis and Mouret, Jean-Baptiste},
  editor     = {Pardalos, Panos M. and Rasskazova, Varvara and Vrahatis, Michael N.},
  year       = {2021},
  volume     = {170},
  pages      = {109--135},
  publisher  = {Springer International Publishing},
  address    = {Cham},
  doi        = {10.1007/978-3-030-66515-9_4},
  urldate    = {2024-01-18},
  isbn       = {978-3-030-66514-2 978-3-030-66515-9}
}

@article{chavira2008WMC,
  title   = {On {{Probabilistic Inference}} by {{Weighted Model Counting}}},
  author  = {Chavira, Mark and Darwiche, Adnan},
  year    = {2008},
  month   = apr,
  journal = {Artificial Intelligence},
  volume  = {172},
  number  = {6},
  pages   = {772--799},
  issn    = {0004-3702},
  doi     = {10.1016/j.artint.2007.11.002},
  urldate = {2021-06-14}
}

@inproceedings{cunnington2021InductiveLearningComplex,
  title     = {Inductive {{Learning}} of {{Complex Knowledge}} from {{Raw Data}}},
  booktitle = {{{AAAI Fall Symposium}}},
  author    = {Cunnington, Daniel and Law, Mark and Lobo, Jorge and Russo, Alessandra},
  year      = {2022},
  volume    = {3332},
  pages     = {1--21},
  publisher = {CEUR Workshop Proceedings},
  address   = {Arlington, Virginia, USA},
  urldate   = {2024-11-28}
}

@misc{cunnington2024NeuroSymbolicLearningAnswer,
  title         = {Neuro-{{Symbolic Learning}} of {{Answer Set Programs}} from {{Raw Data}}},
  author        = {Cunnington, Daniel and Law, Mark and Lobo, Jorge and Russo, Alessandra},
  year          = {2024},
  month         = feb,
  number        = {arXiv:2205.12735},
  eprint        = {2205.12735},
  primaryclass  = {cs},
  publisher     = {arXiv},
  doi           = {10.48550/arXiv.2205.12735},
  urldate       = {2024-06-26},
  archiveprefix = {arXiv}
}

@inproceedings{dai2019Bridging,
  title     = {Bridging {{Machine Learning}} and {{Logical Reasoning}} by {{Abductive Learning}}},
  booktitle = {Advances in {{Neural Information Processing Systems}}},
  author    = {Dai, Wang-Zhou and Xu, Qiuling and Yu, Yang and Zhou, Zhi-Hua},
  editor    = {Wallach, H. and Larochelle, H. and Beygelzimer, A. and {dAlch{\'e}-Buc}, F. and Fox, E. and Garnett, R.},
  year      = {2019},
  volume    = {32},
  pages     = {2811--2822},
  publisher = {Curran Associates, Inc.},
  address   = {Vancouver, Canada}
}

@misc{dai2021MetaAbd,
  title         = {Abductive {{Knowledge Induction From Raw Data}}},
  author        = {Dai, Wang-Zhou and Muggleton, Stephen H.},
  year          = {2021},
  month         = may,
  number        = {arXiv:2010.03514},
  eprint        = {2010.03514},
  primaryclass  = {cs},
  publisher     = {arXiv},
  doi           = {10.48550/arXiv.2010.03514},
  urldate       = {2022-09-29},
  archiveprefix = {arXiv}
}

@misc{daniele2023DSL,
  title         = {Deep {{Symbolic Learning}}: {{Discovering Symbols}} and {{Rules}} from {{Perceptions}}},
  shorttitle    = {Deep {{Symbolic Learning}}},
  author        = {Daniele, Alessandro and Campari, Tommaso and Malhotra, Sagar and Serafini, Luciano},
  year          = {2023},
  month         = apr,
  number        = {arXiv:2208.11561},
  eprint        = {2208.11561},
  primaryclass  = {cs},
  publisher     = {arXiv},
  doi           = {10.48550/arXiv.2208.11561},
  urldate       = {2023-05-12},
  archiveprefix = {arXiv}
}

@inproceedings{darwiche2011SDD,
  title      = {{{SDD}}: {{A New Canonical Representation}} of {{Propositional Knowledge Bases}}},
  shorttitle = {{{SDD}}},
  booktitle  = {Twenty-{{Second International Joint Conference}} on {{Artificial Intelligence}}},
  author     = {Darwiche, Adnan},
  year       = {2011},
  month      = jun,
  pages      = {819--826},
  urldate    = {2021-09-08}
}

@inproceedings{dasgupta2020ExplainableKmeansClustering,
  title      = {Explainable K-Means Clustering: {{Theory}} and Practice},
  shorttitle = {Explainable K-Means Clustering},
  booktitle  = {{{XXAI Workshop}}},
  author     = {Dasgupta, Sanjoy and Frost, Nave and Moshkovitz, Michal and Rashtchian, Cyrus},
  year       = {2020},
  pages      = {1--8},
  address    = {Vienna, Austria},
  urldate    = {2025-11-07}
}

@inproceedings{dessain2023CostExplainabilityAI,
  title      = {Cost of {{Explainability}} in {{AI}}: {{An Example}} with {{Credit Scoring Models}}},
  shorttitle = {Cost of {{Explainability}} in {{AI}}},
  booktitle  = {Explainable {{Artificial Intelligence}}},
  author     = {Dessain, Jean and Bentaleb, Nora and Vinas, Fabien},
  editor     = {Longo, Luca},
  year       = {2023},
  pages      = {498--516},
  publisher  = {Springer Nature Switzerland},
  address    = {Cham},
  doi        = {10.1007/978-3-031-44064-9_26},
  isbn       = {978-3-031-44064-9}
}

@article{diaz-rodriguez2022EXplainable,
  title      = {{{EXplainable Neural-Symbolic Learning}} ({{X-NeSyL}}) Methodology to Fuse Deep Learning Representations with Expert Knowledge Graphs: {{The MonuMAI}} Cultural Heritage Use Case},
  shorttitle = {{{EXplainable Neural-Symbolic Learning}} ({{X-NeSyL}}) Methodology to Fuse Deep Learning Representations with Expert Knowledge Graphs},
  author     = {{D{\'i}az-Rodr{\'i}guez}, Natalia and Lamas, Alberto and Sanchez, Jules and Franchi, Gianni and Donadello, Ivan and Tabik, Siham and Filliat, David and Cruz, Policarpo and Montes, Rosana and Herrera, Francisco},
  year       = {2022},
  month      = mar,
  journal    = {Information Fusion},
  volume     = {79},
  pages      = {58--83},
  issn       = {1566-2535},
  doi        = {10.1016/j.inffus.2021.09.022},
  urldate    = {2022-09-29}
}

@article{evans2018LearningExplanatoryRules,
  title     = {Learning {{Explanatory Rules}} from {{Noisy Data}}},
  author    = {Evans, Richard and Grefenstette, Edward},
  year      = {2018},
  month     = jan,
  journal   = {Journal of Artificial Intelligence Research},
  volume    = {61},
  pages     = {1--64},
  issn      = {1076-9757},
  doi       = {10.1613/jair.5714},
  urldate   = {2024-06-25},
  copyright = {Copyright (c)}
}

@article{garcez2023Neurosymbolic,
  title      = {Neurosymbolic {{AI}}: The 3rd Wave},
  shorttitle = {Neurosymbolic {{AI}}},
  author     = {d'Avila Garcez, Artur and Lamb, Lu{\'i}s C.},
  year       = {2023},
  month      = nov,
  journal    = {Artificial Intelligence Review},
  volume     = {56},
  number     = {11},
  pages      = {12387--12406},
  issn       = {1573-7462},
  doi        = {10.1007/s10462-023-10448-w},
  urldate    = {2024-04-18}
}

@article{garcia2024InterpretablePoliciesPrice,
  title     = {Interpretable {{Policies}} and the {{Price}} of {{Interpretability}} in {{Hypertension Treatment Planning}}},
  author    = {Garcia, Gian-Gabriel P. and Steimle, Lauren N. and Marrero, Wesley J. and Sussman, Jeremy B.},
  year      = {2024},
  month     = jan,
  journal   = {Manufacturing \& Service Operations Management},
  volume    = {26},
  number    = {1},
  pages     = {80--94},
  publisher = {INFORMS},
  issn      = {1523-4614},
  doi       = {10.1287/msom.2021.0373},
  urldate   = {2025-11-07}
}

@inproceedings{gaunt2017Differentiable,
  title     = {Differentiable {{Programs}} with {{Neural Libraries}}},
  booktitle = {Proceedings of the 34th {{International Conference}} on {{Machine Learning}}},
  author    = {Gaunt, Alexander L. and Brockschmidt, Marc and Kushman, Nate and Tarlow, Daniel},
  year      = {2017},
  month     = jul,
  pages     = {1213--1222},
  publisher = {PMLR},
  issn      = {2640-3498},
  urldate   = {2022-06-01}
}

@inproceedings{glorot2010XavierInit,
  title     = {Understanding the Difficulty of Training Deep Feedforward Neural Networks},
  booktitle = {Proceedings of the {{Thirteenth International Conference}} on {{Artificial Intelligence}} and {{Statistics}}},
  author    = {Glorot, Xavier and Bengio, Yoshua},
  year      = {2010},
  month     = mar,
  pages     = {249--256},
  publisher = {{JMLR Workshop and Conference Proceedings}},
  issn      = {1938-7228},
  urldate   = {2024-01-30}
}

@book{hitzler2021NeSyBook,
  title      = {Neuro-{{Symbolic Artificial Intelligence}}: {{The State}} of the {{Art}}},
  shorttitle = {Neuro-{{Symbolic Artificial Intelligence}}},
  editor     = {Hitzler, Pascal and Sarker, Md Kamruzzaman},
  year       = {2021},
  month      = dec,
  series     = {Frontiers in {{Artificial Intelligence}} and {{Applications}}},
  volume     = {342},
  publisher  = {IOS Press},
  doi        = {10.3233/FAIA342},
  urldate    = {2023-04-25},
  isbn       = {978-1-64368-244-0 978-1-64368-245-7}
}

@misc{jang2017Categorical,
  title         = {Categorical {{Reparameterization}} with {{Gumbel-Softmax}}},
  author        = {Jang, Eric and Gu, Shixiang and Poole, Ben},
  year          = {2017},
  month         = aug,
  number        = {arXiv:1611.01144},
  eprint        = {1611.01144},
  primaryclass  = {cs, stat},
  publisher     = {arXiv},
  doi           = {10.48550/arXiv.1611.01144},
  urldate       = {2023-07-07},
  archiveprefix = {arXiv}
}

@incollection{kakas2017Abduction,
  title     = {Abduction},
  booktitle = {Encyclopedia of {{Machine Learning}} and {{Data Mining}}},
  author    = {Kakas, Antonis C.},
  editor    = {Sammut, Claude and Webb, Geoffrey I.},
  year      = {2017},
  edition   = {2nd},
  pages     = {1--8},
  publisher = {Springer US},
  address   = {Boston, MA}
}

@misc{kingma2017AdamOptimizer,
  title         = {Adam: {{A Method}} for {{Stochastic Optimization}}},
  shorttitle    = {Adam},
  author        = {Kingma, Diederik P. and Ba, Jimmy},
  year          = {2017},
  month         = jan,
  number        = {arXiv:1412.6980},
  eprint        = {1412.6980},
  primaryclass  = {cs},
  publisher     = {arXiv},
  doi           = {10.48550/arXiv.1412.6980},
  urldate       = {2023-08-01},
  archiveprefix = {arXiv}
}

@inproceedings{laber2021PriceExplainabilityCluster,
  title     = {On the Price of Explainability for Some Clustering Problems},
  booktitle = {Proceedings of the 38th {{International Conference}} on {{Machine Learning}}},
  author    = {Laber, Eduardo S. and Murtinho, Lucas},
  year      = {2021},
  month     = jul,
  pages     = {5915--5925},
  publisher = {PMLR},
  issn      = {2640-3498},
  urldate   = {2025-11-07}
}

@article{lecun1998GradientBased,
  title   = {Gradient-{{Based Learning Applied}} to {{Document Recognition}}},
  author  = {Lecun, Y. and Bottou, L. and Bengio, Y. and Haffner, P.},
  year    = {1998},
  month   = nov,
  journal = {Proceedings of the IEEE},
  volume  = {86},
  number  = {11},
  pages   = {2278--2324},
  issn    = {1558-2256},
  doi     = {10.1109/5.726791}
}

@article{lenail2019NNSVG,
  title      = {{{NN-SVG}}: {{Publication-Ready Neural Network Architecture Schematics}}},
  shorttitle = {{{NN-SVG}}},
  author     = {LeNail, Alexander},
  year       = {2019},
  month      = jan,
  journal    = {Journal of Open Source Software},
  volume     = {4},
  number     = {33},
  pages      = {747},
  issn       = {2475-9066},
  doi        = {10.21105/joss.00747},
  urldate    = {2024-04-26}
}

@article{liang2021BlackBoxes,
  title      = {Explaining the {{Black-Box Model}}: {{A Survey}} of {{Local Interpretation Methods}} for {{Deep Neural Networks}}},
  shorttitle = {Explaining the {{Black-Box Model}}},
  author     = {Liang, Yu and Li, Siguang and Yan, Chungang and Li, Maozhen and Jiang, Changjun},
  year       = {2021},
  month      = jan,
  journal    = {Neurocomputing},
  volume     = {419},
  pages      = {168--182},
  issn       = {0925-2312},
  doi        = {10.1016/j.neucom.2020.08.011},
  urldate    = {2024-04-15}
}

@misc{maddison2017Concrete,
  title         = {The {{Concrete Distribution}}: {{A Continuous Relaxation}} of {{Discrete Random Variables}}},
  shorttitle    = {The {{Concrete Distribution}}},
  author        = {Maddison, Chris J. and Mnih, Andriy and Teh, Yee Whye},
  year          = {2017},
  month         = mar,
  number        = {arXiv:1611.00712},
  eprint        = {1611.00712},
  primaryclass  = {cs, stat},
  publisher     = {arXiv},
  doi           = {10.48550/arXiv.1611.00712},
  urldate       = {2023-07-07},
  archiveprefix = {arXiv}
}

@misc{mao2019NSCL,
  title         = {The {{Neuro-Symbolic Concept Learner}}: {{Interpreting Scenes}}, {{Words}}, and {{Sentences From Natural Supervision}}},
  shorttitle    = {The {{Neuro-Symbolic Concept Learner}}},
  author        = {Mao, Jiayuan and Gan, Chuang and Kohli, Pushmeet and Tenenbaum, Joshua B. and Wu, Jiajun},
  year          = {2019},
  month         = apr,
  number        = {arXiv:1904.12584},
  eprint        = {1904.12584},
  primaryclass  = {cs},
  publisher     = {arXiv},
  doi           = {10.48550/arXiv.1904.12584},
  urldate       = {2024-05-15},
  archiveprefix = {arXiv}
}

@inproceedings{markos2022ProxyCoaches,
  title     = {Machine {{Coaching}} with {{Proxy Coaches}}},
  booktitle = {Proceedings of the 1st {{Workshop}} on {{Argumentation}} \& {{Machine Learning}}},
  author    = {Markos, Vassilis and Thoma, Marios and Michael, Loizos},
  editor    = {Kuhlmann, Isabelle and Mumford, Jack and Sarkadi, Stefan},
  year      = {2022},
  series    = {{{CEUR Workshop Proceedings}}},
  volume    = {3208},
  pages     = {45--64},
  publisher = {CEUR},
  address   = {Cardiff, Wales},
  issn      = {1613-0073},
  urldate   = {2023-03-03}
}

@inproceedings{markos2022Prudens,
  title      = {Prudens: {{An Argumentation-Based Language}} for {{Cognitive Assistants}}},
  shorttitle = {Prudens},
  booktitle  = {Rules and {{Reasoning}}},
  author     = {Markos, Vassilis and Michael, Loizos},
  editor     = {Governatori, Guido and Turhan, Anni-Yasmin},
  year       = {2022},
  series     = {Lecture {{Notes}} in {{Computer Science}}},
  pages      = {296--304},
  publisher  = {Springer International Publishing},
  address    = {Cham},
  doi        = {10.1007/978-3-031-21541-4_19},
  isbn       = {978-3-031-21541-4}
}

@article{marra2024Statistical,
  title     = {From {{Statistical Relational}} to {{Neuro-Symbolic Artificial Intelligence}}},
  author    = {Marra, Giuseppe},
  year      = {2024},
  month     = mar,
  journal   = {Proceedings of the AAAI Conference on Artificial Intelligence},
  volume    = {38},
  number    = {20},
  pages     = {22678--22678},
  issn      = {2374-3468},
  doi       = {10.1609/aaai.v38i20.30294},
  urldate   = {2024-04-26},
  copyright = {Copyright (c) 2024 Association for the Advancement of Artificial Intelligence}
}

@inproceedings{mccarthy1998Elaboration,
  title     = {Elaboration {{Tolerance}}},
  booktitle = {Common {{Sense}} 98},
  author    = {McCarthy, John},
  year      = {1998},
  volume    = {98},
  pages     = {2},
  address   = {London, U.K.}
}

@inproceedings{michael2019MachineCoaching,
  ids       = {michael},
  title     = {Machine {{Coaching}}},
  booktitle = {{{IJCAI}} 2019 {{Workshop}} on {{Explainable Artificial Intelligence}}},
  author    = {Michael, Loizos},
  year      = {2019},
  month     = aug,
  pages     = {80--86},
  address   = {Macau, China}
}

@incollection{michael2023Chapter11Autodidactic,
  title     = {Chapter 11. {{Autodidactic}} and {{Coachable Neural Architectures}}},
  booktitle = {Compendium of {{Neurosymbolic Artificial Intelligence}}},
  author    = {Michael, Loizos},
  editor    = {Hitzler, Pascal and Sarker, Md Kamruzzaman and Eberhart, Aaron},
  year      = {2023},
  month     = jul,
  series    = {Frontiers in {{Artificial Intelligence}} and {{Applications}}},
  volume    = {369},
  pages     = {235--248},
  publisher = {IOS Press},
  doi       = {10.3233/FAIA230143},
  urldate   = {2025-05-12},
  isbn      = {978-1-64368-406-2 978-1-64368-407-9}
}

@misc{mouret2015MappingElites,
  title         = {Illuminating Search Spaces by Mapping Elites},
  author        = {Mouret, Jean-Baptiste and Clune, Jeff},
  year          = {2015},
  month         = apr,
  number        = {arXiv:1504.04909},
  eprint        = {1504.04909},
  primaryclass  = {cs, q-bio},
  publisher     = {arXiv},
  doi           = {10.48550/arXiv.1504.04909},
  urldate       = {2024-01-22},
  archiveprefix = {arXiv}
}

@misc{pryor2023NeuPSL,
  title         = {{{NeuPSL}}: {{Neural Probabilistic Soft Logic}}},
  shorttitle    = {{{NeuPSL}}},
  author        = {Pryor, Connor and Dickens, Charles and Augustine, Eriq and Albalak, Alon and Wang, William and Getoor, Lise},
  year          = {2023},
  month         = may,
  number        = {arXiv:2205.14268},
  eprint        = {2205.14268},
  primaryclass  = {cs},
  publisher     = {arXiv},
  doi           = {10.48550/arXiv.2205.14268},
  urldate       = {2023-12-15},
  archiveprefix = {arXiv}
}

@misc{serafini2016LogicTensorNets,
  title         = {Logic {{Tensor Networks}}: {{Deep Learning}} and {{Logical Reasoning}} from {{Data}} and {{Knowledge}}},
  shorttitle    = {Logic {{Tensor Networks}}},
  author        = {Serafini, Luciano and d'Avila Garcez, Artur},
  year          = {2016},
  month         = jul,
  number        = {arXiv:1606.04422},
  eprint        = {1606.04422},
  primaryclass  = {cs},
  publisher     = {arXiv},
  doi           = {10.48550/arXiv.1606.04422},
  urldate       = {2024-01-24},
  archiveprefix = {arXiv}
}

@inproceedings{sha2023NeSyPredicateInvention,
  title     = {Neural-{{Symbolic Predicate Invention}}: {{Learning Relational Concepts}} from {{Visual Scenes}}},
  booktitle = {{{NeSy}} 2023, 17th {{International Workshop}} on {{Neural-Symbolic Learning}} and {{Reasoning}}},
  author    = {Sha, Jingyuan and Shindo, Hikaru and Kersting, Kristian and Dhami, Devendra Singh},
  year      = {2023},
  volume    = {3432},
  pages     = {1--15},
  publisher = {CEUR Workshop Proceedings},
  address   = {Certosa di Pontignano, Siena, Italy}
}

@article{trinh2024AlphaGeometry,
  title     = {Solving {{Olympiad Geometry Without Human Demonstrations}}},
  author    = {Trinh, Trieu H. and Wu, Yuhuai and Le, Quoc V. and He, He and Luong, Thang},
  year      = {2024},
  month     = jan,
  journal   = {Nature},
  volume    = {625},
  number    = {7995},
  pages     = {476--482},
  publisher = {Nature Publishing Group},
  issn      = {1476-4687},
  doi       = {10.1038/s41586-023-06747-5},
  urldate   = {2024-01-23},
  copyright = {2024 The Author(s)}
}

@article{tsamoura2021Neurolog,
  title      = {Neural-{{Symbolic Integration}}: {{A Compositional Perspective}}},
  shorttitle = {Neural-{{Symbolic Integration}}},
  author     = {Tsamoura, Efthymia and Hospedales, Timothy and Michael, Loizos},
  year       = {2021},
  journal    = {Proceedings of the AAAI Conference on Artificial Intelligence},
  volume     = {35},
  number     = {6},
  pages      = {5051--5060},
  issn       = {2374-3468},
  urldate    = {2022-03-17}
}

@article{valiant2009Evolvability,
  title   = {Evolvability},
  author  = {Valiant, Leslie G.},
  year    = {2009},
  journal = {Journal of the ACM},
  volume  = {56},
  number  = {1},
  pages   = {1--21},
  issn    = {00045411},
  doi     = {10.1145/1462153.1462156}
}

@article{valiantPAC1984,
  title   = {A {{Theory}} of the {{Learnable}}},
  author  = {Valiant, Leslie G.},
  year    = {1984},
  journal = {Communications of the ACM},
  volume  = {27},
  number  = {11},
  pages   = {1134--1142},
  issn    = {00010782},
  doi     = {10.1145/1968.1972}
}

@article{vassiliades2018VoronoiPhenotypic,
  title   = {Using {{Centroidal Voronoi Tessellations}} to {{Scale Up}} the {{Multidimensional Archive}} of {{Phenotypic Elites Algorithm}}},
  author  = {Vassiliades, Vassilis and Chatzilygeroudis, Konstantinos and Mouret, Jean-Baptiste},
  year    = {2018},
  month   = aug,
  journal = {IEEE Transactions on Evolutionary Computation},
  volume  = {22},
  number  = {4},
  pages   = {623--630},
  issn    = {1941-0026},
  doi     = {10.1109/TEVC.2017.2735550}
}

@inproceedings{vermeulen2023Experimental,
  title     = {An {{Experimental Overview}} of~{{Neural-Symbolic Systems}}},
  booktitle = {Inductive {{Logic Programming}}},
  author    = {Vermeulen, Arne and Manhaeve, Robin and Marra, Giuseppe},
  editor    = {Bellodi, Elena and Lisi, Francesca Alessandra and Zese, Riccardo},
  year      = {2023},
  series    = {Lecture {{Notes}} in {{Computer Science}}},
  pages     = {124--138},
  publisher = {Springer Nature Switzerland},
  address   = {Cham},
  doi       = {10.1007/978-3-031-49299-0_9},
  isbn      = {978-3-031-49299-0}
}

@misc{wannesm2024PySDD,
  title   = {{{PySDD}}},
  author  = {{wannesm}},
  year    = {2024},
  month   = mar,
  urldate = {2024-04-26}
}

@inproceedings{xu2018SemanticLoss,
  title     = {A {{Semantic Loss Function}} for {{Deep Learning}} with {{Symbolic Knowledge}}},
  booktitle = {Proceedings of the 35th {{International Conference}} on {{Machine Learning}}},
  author    = {Xu, Jingyi and Zhang, Zilu and Friedman, Tal and Liang, Yitao and {Van den Broeck}, Guy},
  year      = {2018},
  month     = jul,
  pages     = {5502--5511},
  publisher = {PMLR},
  address   = {Stockholm, Sweden},
  issn      = {2640-3498},
  urldate   = {2022-07-12}
}

@misc{zhong2023ChatABLAbductiveLearning,
  title         = {{{ChatABL}}: {{Abductive Learning}} via {{Natural Language Interaction}} with {{ChatGPT}}},
  shorttitle    = {{{ChatABL}}},
  author        = {Zhong, Tianyang and Wei, Yaonai and Yang, Li and Wu, Zihao and Liu, Zhengliang and Wei, Xiaozheng and Li, Wenjun and Yao, Junjie and Ma, Chong and Li, Xiang and Zhu, Dajiang and Jiang, Xi and Han, Junwei and Shen, Dinggang and Liu, Tianming and Zhang, Tuo},
  year          = {2023},
  month         = apr,
  number        = {arXiv:2304.11107},
  eprint        = {2304.11107},
  primaryclass  = {cs},
  publisher     = {arXiv},
  doi           = {10.48550/arXiv.2304.11107},
  urldate       = {2024-06-25},
  archiveprefix = {arXiv}
}
